\documentclass{article} %
\usepackage[final]{colm2026_conference}

\providecommand{\danqi}[1]{{\protect\color{cyan}{[Danqi: #1]}}}

\providecommand{\howard}[1]{{\protect\color{purple}{[Howard: #1]}}}

\definecolor{mutedgreen}{HTML}{2E8B57}

\providecommand{\danqi}[1]{{\protect\color{cyan}{}}}
\providecommand{\howard}[1]{{\protect\color{purple}{}}}

\newcommand\ti[1]{\textit{#1}}

\newcommand\tf[1]{\textbf{#1}}

\renewcommand{\paragraph}[1]{\vspace{0.2cm}\noindent\textbf{#1}}

\newcommand{\react}{\textsc{ReAct}}
\newcommand{\searcho}{\textsc{Search-o1}}
\newcommand{\gptr}{\textsc{GPT-R}}
\newcommand{\hfodr}{\textsc{HF-ODR}}
\newcommand{\gptrfull}{GPT-Researcher}
\newcommand{\hfodrfull}{HuggingFace OpenDeepResearch}

\newcommand{\glm}{GLM-4.7 Flash}
\newcommand{\gptoss}{GPT-OSS-120B}
\newcommand{\tongyidr}{Tongyi-DeepResearch-30B}

\newcommand{\ours}{\textsc{Slim}}

\newcommand{\oursfull}{\textsc{Slim} (\textbf{S}imple \tf{\textsc{L}}ightweight \textbf{I}nformation \textbf{M}anagement)}

\newcommand{\headercolor}{\rowcolor{gray!15}}

\usepackage{amsmath}
\DeclareMathOperator*{\argmax}{arg\,max}
\usepackage{microtype}
\usepackage{hyperref}
\usepackage{url}
\usepackage{booktabs}
\usepackage{multirow}
\usepackage[table]{xcolor}

\usepackage{nicematrix}
\usepackage{colortbl}
\usepackage{textcomp} 
\usepackage{hyperref}
\usepackage{cleveref}
\usepackage{tabularx}
\usepackage{wrapfig}
\usepackage[ruled,vlined]{algorithm2e}
\usepackage[textsize=tiny]{todonotes}
\usepackage{fancyvrb}
\usepackage{lineno}

\definecolor{darkblue}{rgb}{0, 0, 0.5}
\hypersetup{colorlinks=true, citecolor=darkblue, linkcolor=darkblue, urlcolor=darkblue}

\title{Lost in the Maze: Overcoming Context Limitations in Long-Horizon Agentic Search}

\author{Howard Yen$^{\hspace{.1em}{\color{orange}\boldsymbol{p}}}$\thanks{Work done as an intern at Samaya AI.}$\;$ 
Yoonsang Lee$^{\hspace{.1em}{\color{orange}\boldsymbol{p}}}$  $\;$
Ashwin Paranjape$^{\hspace{.1em}{\color{blue}\boldsymbol{s}}}$ $\;$
Mengzhou Xia$^{\hspace{.1em}{\color{orange}\boldsymbol{p}}}$ $\;$\\
\tf{Thejas Venkatesh}$^{\hspace{.1em}{\color{blue}\boldsymbol{s}}}$ $\;$
~\tf{Jack Hessel}$^{\hspace{.1em}{\color{blue}\boldsymbol{s}}}$ $\;$
\tf{Danqi Chen}$^{\hspace{.1em}{\color{orange}\boldsymbol{p}}}$ $\;$
\tf{Yuhao Zhang}$^{\hspace{.1em}{\color{blue}\boldsymbol{s}}}$\\
$^{\hspace{.1em}{\color{orange}\boldsymbol{p}}}$Princeton Language and Intelligence, Princeton University $^{\hspace{.1em}{\color{blue}\boldsymbol{s}}}$Samaya AI\\
\texttt{hyen@cs.princeton.edu}
}

\begin{document}

\ifcolmsubmission
\linenumbers
\fi

\maketitle

\begin{abstract}

Long-horizon agentic search requires iteratively exploring the web over long trajectories and synthesizing information across many sources, enabling powerful applications like deep research systems.
In this work, we show that popular agentic search frameworks struggle to scale to long trajectories primarily due to context limitations—they accumulate long, noisy content, hit context window and tool budgets, or stop early. 
We therefore introduce \oursfull, a simple framework that separates retrieval into distinct search and browse tools, and periodically summarizes the trajectory, keeping context concise while enabling longer, more focused searches. 
Across a wide range of long-horizon tasks, \ours{} achieves comparable performance at substantially lower cost and far fewer tool calls than strong open-source frameworks with both proprietary and open-weight models, including RL-trained models for deep research.
Specifically, with o3 as the base model, \ours{} achieves 56\% on BrowseComp and 33\% on HLE, outperforming all open-source frameworks by 8 and 6 absolute points, respectively, while incurring 4--6x fewer tool calls.
With \glm, \ours{} achieves 10 points improvement over the next best open-source framework, \searcho, on BrowseComp using a third of the cost.
To systematically understand failure modes in long-horizon agentic search, we develop an automated fine-grained trajectory analysis pipeline and error taxonomy, and find that \ours{} exhibits significantly fewer hallucinations than prior systems.
We hope our analysis framework and simple tool design inform future long-horizon agents\footnote{Code is available at \url{https://github.com/howard-yen/SLIM}}.

\end{abstract}

\section{Introduction}

Long-horizon agentic search involves performing searches over long trajectories and reasoning over many sources, and requires powerful systems that can explore diverse sources and leverage tools effectively.
The ability to reason over long trajectories serves as the foundation for exciting applications such as deep research \citep{OpenAIDeepResearch2025,GoogleGeminiDeepResearch2025,xAI_Grok32025}.
Due to its immense potential in solving complex tasks, long-horizon systems have been a key focus in the community, eliciting the development of many proprietary and open-source frameworks.
Among open-source systems, HuggingFace Open Deep Research \citep{huggingface2025opendeepresearch} and GPT Researcher \citep{Elovic_gpt-researcher_2023} opt for complex multi-agent orchestration while \searcho{} \citep{li2025searcho1agenticsearchenhancedlarge} uses a single agent.
However, despite the numerous approaches, they still fail in complex long-trajectory settings, and there are no systematic approaches to analyze their trajectories and identify the failure modes.

In this work, we first analyze existing frameworks by examining their trajectory outcomes on 
BrowseComp \citep{wei2025browsecompsimplechallengingbenchmark}, a challenging long-horizon agentic search benchmark.
Our analysis shows that these frameworks still struggle with long-trajectory tasks, failing on more than 50\% of the samples---most of the failures are due to hitting the context window limit, running out of tool budget, or stopping too early.

\begin{figure}[t!]
    \centering
    \includegraphics[width=0.95\linewidth]{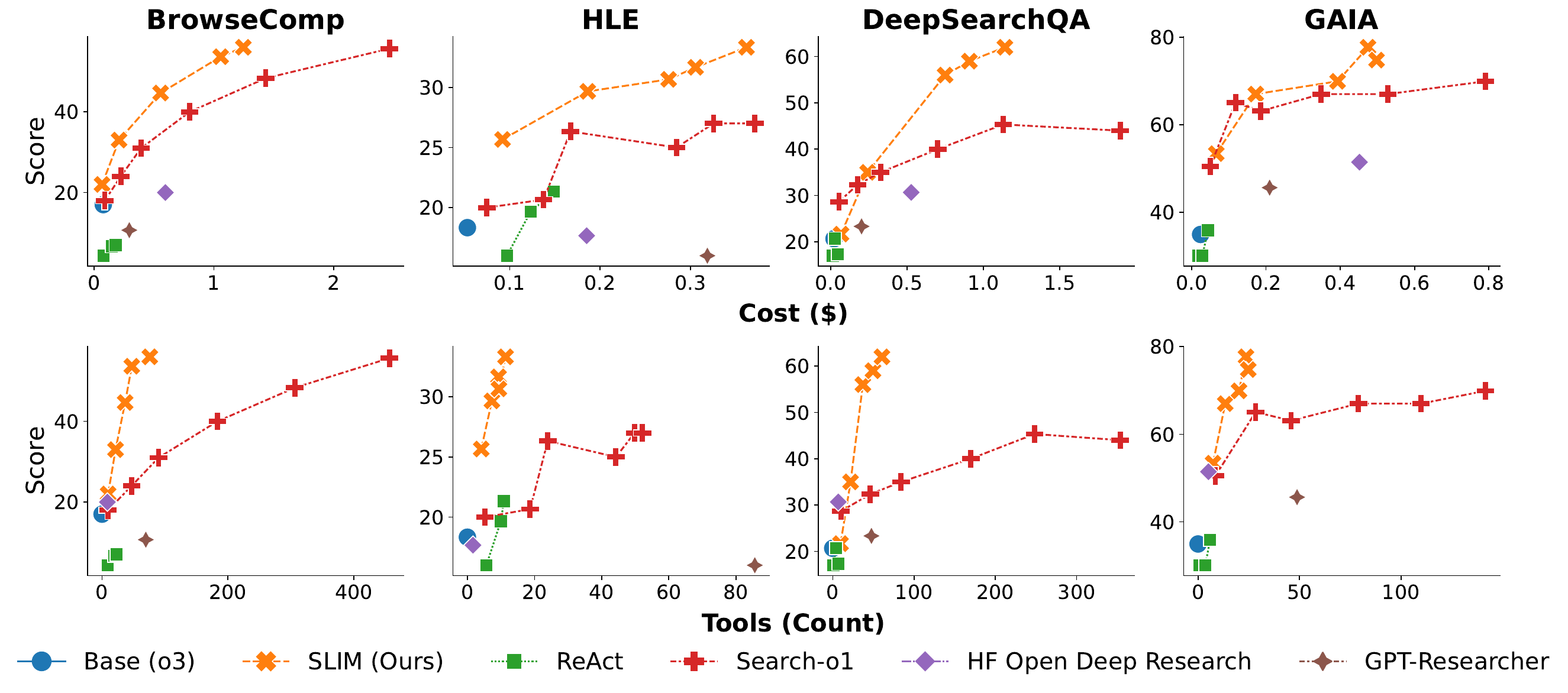}
    \vspace{-4.0pt}
    \caption{
        With o3 as the base model, \ours{} achieves better performance than existing frameworks on across all datasets while using more than 4-6x fewer tool calls and lower overall costs, which account for LLM token usage and tool costs. 
    }
    \vspace{-14.0pt}
    \label{fig:main}
\end{figure}

We attribute these failure modes to poor context management that can fill the context window with noisy information that derails long search trajectories.
The limited context restricts the number of turns in each trajectory, resulting in incomplete information gathering.
To overcome these limitations, we design \oursfull{}, a framework with three simple yet powerful components---search, browse, and summarization---that effectively manage the context size of long-horizon systems.
The simple tool design allows LLMs to interleave searching for diverse information and browsing promising pages without spending unnecessary tool calls on noisy results. 
Furthermore, the summarization module acts as a general-purpose context manager that can reduce long trajectories into more condensed summaries.
These design choices combine to enable scaling to longer trajectories while maintaining a concise context and reduced tool costs.
Under a comparable cost budget, with o3 as the base model, \ours{} significantly outperforms the previous best open-source frameworks across all benchmarks, while requiring only 15-25\% of the tool calls (\Cref{fig:main}).

Finally, we develop an automated trajectory-level analysis pipeline that provides fine-grained insights into long-horizon frameworks.
To characterize mistakes made by these systems, we develop an error taxonomy identifying common failure modes.
Our analysis reveals that \ours{}'s advantage stems from its robustness to failure modes such as hallucinations and unfocused and generic searches.
We hope our analysis pipeline, error taxonomy, and careful design choices in \ours{} can serve as a foundation for understanding and improving long-horizon agentic search systems.

\section{Preliminaries: Long-Horizon Agentic Search}

Previous search tasks can be mostly solved with static retrieval-augmented generation (RAG) systems that leverage at most a few retrieval steps \citep{lewis2020rag,izacard2023atlas,shi-etal-2024-replug}, and do not showcase the challenges of realistic, long-horizon agentic search settings.
In contrast, we study long-horizon tasks with complex queries that require extensive searches to gather the necessary information and reasoning over different sources to synthesize the answer.
In this section, we formalize the task, describe the datasets for studying long-horizon agentic search, and review some previous long-horizon systems.

\subsection{Task Formulation}

We formalize long-horizon agentic search tasks as follows: given a query $q$, a corpus of documents $\mathcal{D}$, the system needs to perform a sequence of tool calls 
to find relevant information from $\mathcal{D}$ and output a final answer $o$, which is checked against the annotated groundtruth answer $a$.
A critical component of the system is the design of its tools and how it interacts with the corpus; each tool is a function $\mathcal{T}_i(x) \rightarrow y$ that maps arbitrary system-generated inputs $x$ to arbitrary outputs $y$.

Furthermore, agentic systems are often controlled by a tool budget $T$, the maximum number of tool calls they are allowed to use in any trajectory.
The tool budget $T$ also corresponds to the maximum number of turns in a trajectory, as each turn corresponds to one tool call\footnote{Some architectures, such as the CodeAgent \citep{wang2024codeagent} used in \hfodr, allow for parallel tool calls in one step, but we found that the models we tested do not use this capability.}.
Thus, how to manage the input context to the underlying LLM across many tool uses and turns is another critical design choice in long-horizon systems.
Finally, the final step where the system outputs its final answer does not count towards the tool budget.

In long-horizon agentic search settings, the web is most often used as the corpus $\mathcal{D}$ due to the diversity and complexity of the queries, and each document $d_i = (u_i, t_i, c_i)$ comprises a URL, title, and content.
In practice, long-horizon systems typically use search engines $\mathcal{R}(q) \rightarrow \{(u_i, t_i)\}_1^n$ to obtain a list of $n$ web pages with their titles and URLs most relevant to the search query $q$.
Furthermore, a scraping operation $\mathcal{C}(u_i) \rightarrow c_i$ is necessary to obtain the full content of any URL as search engines only provide a list of URLs, but scraping is slow and noisy in practice.

In traditional QA settings, since the retrieval tool only needs to be called once due to the simplicity of the queries and the small size of the corpus (i.e., Wikipedia), retrieval returns the full list of documents and their contents $\mathcal{R}_{\texttt{wiki}}(q) \rightarrow \{(t_i, c_i)\}_1^n$.
As a result, many long-horizon systems follow a similar design, where the retrieval tool is a single search engine call followed by scraping all returned URLs.
However, the complexity of long-horizon agentic search requires many tool calls to gather the necessary information \citep{li2025searcho1agenticsearchenhancedlarge,jin2025searchr}.
As we demonstrate empirically later, this naive tool design leads to severe context limitations, where the system is overwhelmed by long, noisy content, motivating the design of more efficient tool interfaces for long-horizon systems.

\subsection{Datasets}

\begin{figure}[t!]
    \centering
    \includegraphics[width=0.95\linewidth]{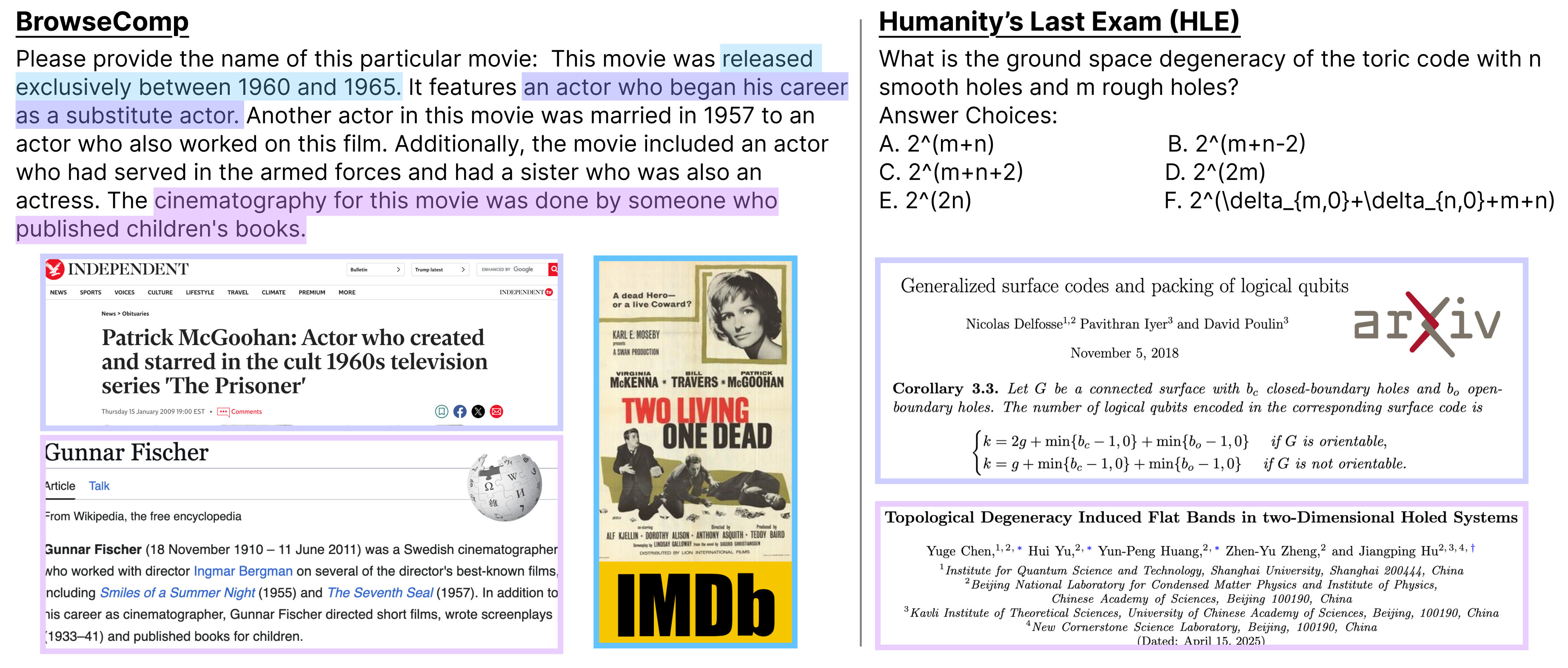}
    \vspace{-2.0pt}
    \caption{
        Example queries and their relevant documents for BrowseComp \citep{wei2025browsecompsimplechallengingbenchmark} and HLE \citep{phan2025humanitysexam}. 
    }
    \label{fig:data_example}
\end{figure}

We select five diverse datasets with naturally difficult queries that require long-trajectory searches and verifiable answers, which ensures the reliability of subsequent analyses.
For evaluation, we sample a random subset of 300 instances from each dataset due to the high costs of running long-horizon systems.\footnote{For GAIA, we evaluate on the subset of 103 text only queries, following \citet{liu2025webexplorerexploreevolvetraining}.}. 
Examples are shown in \Cref{fig:data_example}.

BrowseComp \citep{wei2025browsecompsimplechallengingbenchmark} and GAIA \citep{mialon2023gaiabenchmarkgeneralai} contain challenging queries that test the ability to exhaustively search the web over long trajectories.
Humanity's Last Exam \citep{phan2025humanitysexam} tests across multiple domains and often requires domain-specific knowledge and applying web information to reasoning-heavy problems.
DeepSearchQA \citep{gupta2026deepsearchqabridgingcomprehensivenessgap} and HealthBench \citep{arora2025healthbenchevaluatinglargelanguage} evaluate long-form generation capabilities across board domains.

\begin{table}[!th]
    \centering
    \small
    \resizebox{0.98\linewidth}{!}{
        \begin{tabular}{lrrrrr}
            \toprule
            Framework & Architecture & \# Tools & Tools & Input to LLM Context & Summarization \\
            \midrule
            \react & Single-agent & 1 & Retrieval & All search results & - \\
            \searcho & Single-agent & 1 & Retrieval & All search results & Retrieved content \\
            \hfodr & Multi-agent & 11 & Search, Browse, Python, ... & Selected search results & Search agent result \\
            \gptr & Multi-agent & 1 & Retrieval & All search results & Retrieved content \\
            \midrule
            \ours~(ours) & Single-agent & 2 & Search, Browse & Selected search results & Task trajectory\\
            \bottomrule
        \end{tabular}
    }
    \caption{
        Comparison of \ours{} with existing frameworks. In contrast to single-agent works that bundle search and browsing search results into \emph{one} retrieval tool, we separate it into two distinct tools. 
    }
    \label{tab:framework_comparison}
\end{table}

\subsection{Existing Approaches}

We briefly describe some popular approaches to agentic search, ranging from simple single-LLM frameworks to complex multi-agent systems.
We summarize the differences between these frameworks in \Cref{tab:framework_comparison}; more details are in \S\ref{app:existing}. For \react{} and \searcho, we vary the tool budget, and for \hfodr{} and \gptr, we follow the default settings.

\textbf{\react} \citep{yao2023react} is a simple framework that allows an LLM agent to alternate between thinking and acting, allowing tool calling across many turns.
Following the original work, LLM uses a single retrieval tool---given a query, the tool returns a list of top 10 results along with their web contents.
All results are then concatenated to the agent's context for subsequent steps.
\textbf{\searcho} \citep{li2025searcho1agenticsearchenhancedlarge} builds upon \react{} with an additional ``reason-in-document'' step, where an LLM summarizes the search results and their contents before appending the results to the agent's input context.
Although summarization reduces context length for the main LLM compared to \react, this approach still uses many scraping operations in each search step (one for each result), and summarization incurs a large amount of token usage.
\textbf{\hfodrfull} \citep[\hfodr;][]{huggingface2025opendeepresearch} leverages a hierarchical structure consisting of a manager agent and a search agent, which iteratively interacts with a search engine, a browser, and other tools (detailed in \S\ref{tools:hfodr}).
\textbf{\gptrfull} \citep[\gptr;][]{Elovic_gpt-researcher_2023} is a complex multi-agent system where each agent has distinct roles:
research conductor, report generator, context manager, and source curator.
The system uses a deep researcher agent that acts as a search tree node, spawning multiple children nodes with these same components.

\section{Failure Modes of Existing Approaches}

Despite recent progress, we still know little about how individual components in these systems perform, or fail. To study behavior on long-horizon tasks, we focus on BrowseComp, which naturally induces extended, multi-step search trajectories.
For this task, the final outcome can reveal the overall performance of each framework as well as its relationship with the context window limitation and tool budget constraints.
For this analysis, we let the framework run up to a fixed number of turns and output an answer. 
We categorize the final outcome in \Cref{tab:outcome_descriptions}.

\begin{table}[th]

    \centering 
    \small
    \begin{tabularx}{\textwidth}{@{}lX@{}}
        \toprule
        \textbf{Outcome} & \textbf{Description} \\ 
        \midrule
        Correct          & The system outputs the correct answer                                                                                                                                             \\
        Exceed context   & The system exceeds LLM's context window, falling back to not using any tools                                                                              \\
        Exceed budget    & The system exceeds the tool calling or iteration budget                                                                                       \\
        Early stopping   & The system outputs an incorrect answer before reaching the iteration budget                                                                                       \\
        No tool used & A special case of early stopping where the system does not use any tools                                                                                       \\
        Misc. error      & Due to uncontrollable factors (e.g., API content filters) the system outputs an error %

        \\ 
        \bottomrule
    \end{tabularx}

    \caption{Categorization of different search outcomes and their descriptions.}
    \label{tab:outcome_descriptions}
\end{table}

\begin{figure}[t!]
    \centering
    \includegraphics[width=0.95\linewidth]{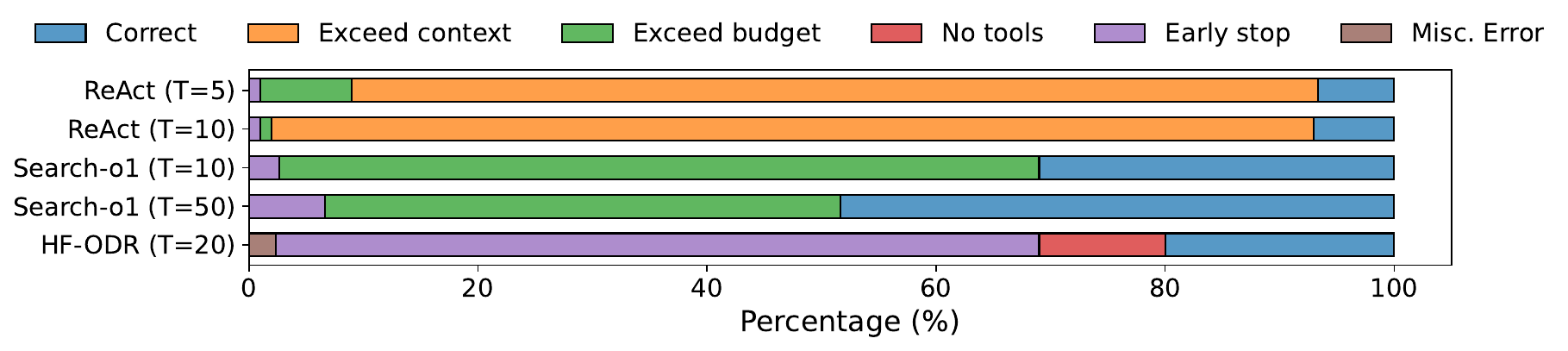}
    \vspace{-3.0pt}
    \caption{
        Each framework exhibits distinct outcome trends---\react{} predominantly runs out of context window, while \searcho{} is often limited by the tool budget (T).
        We exclude \gptr{} due to its predefined workflow---the outcome can only be either correct or incorrect. 
    }
    \label{fig:analysis_outcome}
\end{figure}

For this analysis, we consider different tool budgets for \react{} and \searcho, and use the default 20 turns for \hfodr.
We observe that context window limitations and tool budgets are the main bottlenecks for existing approaches in \Cref{fig:analysis_outcome}, and each framework exhibits distinct patterns.

Specifically, \react{} often hits the context window limit over a long trajectory due to the large amount of text returned by each search call.
As a result, it cannot effectively scale to long trajectories and make full use of its tool budgets.
\searcho{} failure cases are almost entirely due to exceeding the tool budget, which suggests increasing the tool budget may potentially lead to better performance.
However, such an increase is non-trivial without incurring a significant amount of cost---each retrieval step in \searcho{} involves scraping all search results, even though only a fraction of these results are relevant, leading to a large amount of LLM token consumption during the summarization step.

Finally, we observe that \hfodr{} often prematurely terminates because the manager agent cannot leverage its search agent across multiple steps.
Furthermore, \hfodr{} is the only framework that do not use any tools in a significant percentage of the trajectories (10\%), suggesting that complex prompt-engineered workflows may be prone to reducing the tool calling capabilities of the base model.
The root cause of these failure modes is poor context management---exceeding context and tool budgets, or stopping too early. 
In the next section, we explore how to substantially improve agentic search  through better context management. %

\section{Our Framework: \ours}

\begin{figure}[t!]
    \centering
    \includegraphics[width=0.98\linewidth]{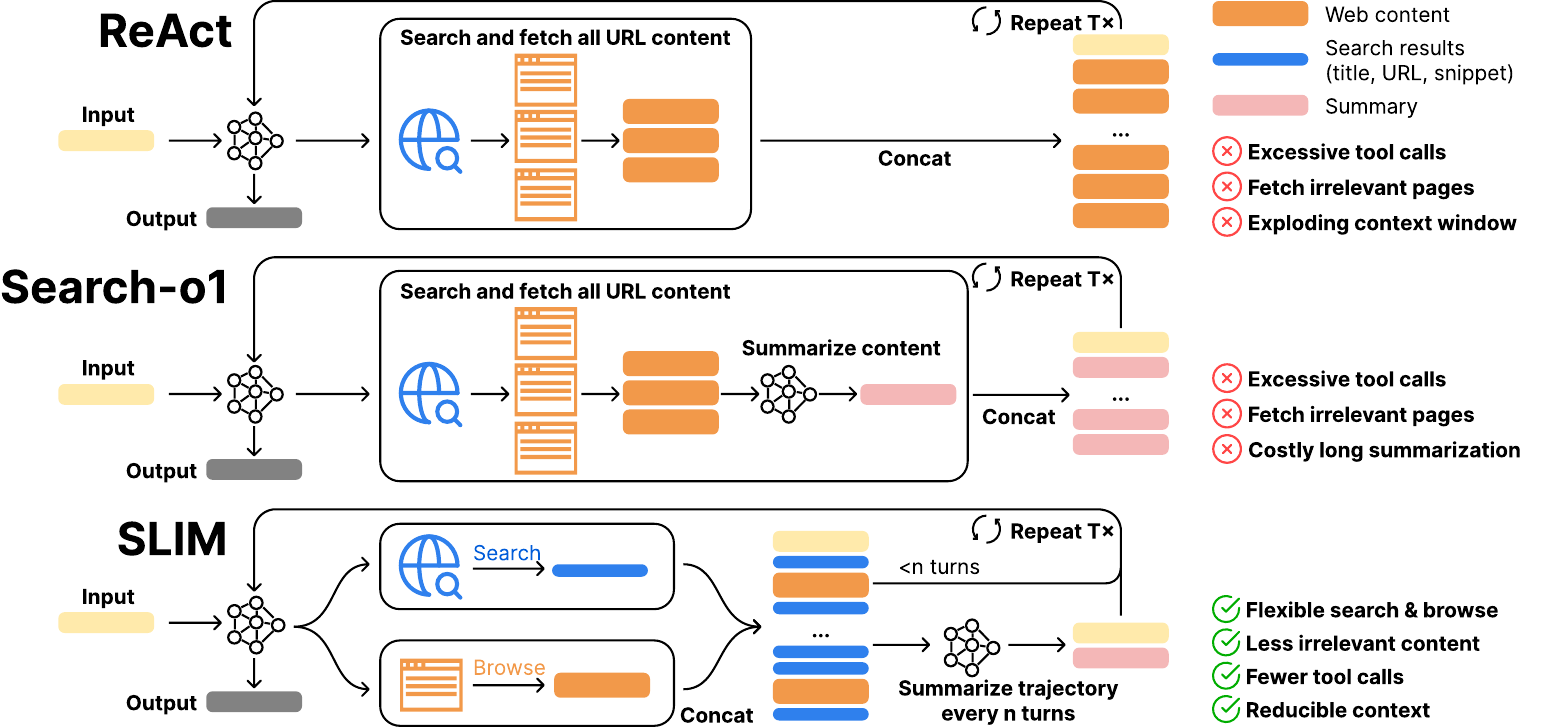}
    \vspace{-2.0pt}
    \caption{
        Compared to \react{} and \searcho, the cooperation between search, browse, and summarization modules allows \ours{} to accumulate shorter contexts and less noisy information after exploring the same amount of searches.
    }
    \label{fig:framework}
\end{figure}

A key takeaway from our analysis is that long-trajectory tasks require \textbf{scaling up the number of turns and tool calls while keeping the context concise to avoid hitting the context window limit}.
Specifically, search results are often noisy and irrelevant to the answer, so filling up the context with content from all search results can lead to noisy context and unnecessary tool costs.
Motivated by these observations, we introduce \oursfull{} with two key principles: (1) using simple and flexible tools for LLMs to interact with, and (2) minimizing the amount of noisy information presented to the model and keeping the context concise during exploration.
An overview of \ours{} in comparison to existing frameworks is shown in \Cref{fig:framework}.

Concretely, \ours{} adopts three simple yet powerful components---search, browse, and summarization---to effectively manage the context and scale the number of turns.

\paragraph{Search tool $\mathcal{R}$.} 
\ours{} uses a search tool that only returns the top $k$ search results from a search engine, where each search result consists of a title, a URL, and a short snippet of its content.
This is much more efficient than previous frameworks that return the full content for all search results, and relies on the main LLM to discern relevant context.

\paragraph{Browse tool $\mathcal{B}$.}
Our browse tool is designed to complement the search tool by allowing the LLM to dig deeper into promising search results.
Specifically, the browse tool $\mathcal{B}(u, q) \rightarrow \max_{c_{i} \in c} \texttt{sim}(c_i, q)$ returns the most relevant section of the content $c$ from the URL $u$ to the query $q$, which enables the LLM to select the most relevant search result and choose a subset of the content that best matches the specific information it is looking for.
As a result, our browse tool is significantly more efficient and cheaper than previous frameworks that exhaustively browse all search results in terms of both the scraping operations and the amount of new tokens introduced to the context.

\paragraph{Summarization module $\mathcal{S}$.}
Despite the brevity of each tool response, agent context inevitably grows as it explores over a long horizon of searches.
To maintain a concise context while retaining the effective exploration history, we introduce a summarization module that periodically compresses the LLM context.
We find a simple heuristic sufficient: we summarize the entire conversation history after every $n$ turns of tool calls and replace the trajectory with the summary.
This crucially differs from previous works where summarization is solely applied to search results at each turn.

Finally, we combine these components into a single framework by allowing the underlying LLM to call either the search or the browse tools at every turn.
Then, the summarization module compresses the entire conversation every $n$ turns to reduce the amount of noise.
Our implementation uses Google\footnote{\url{https://serper.dev/}} as the search tool, crawl4ai\footnote{\url{https://github.com/unclecode/crawl4ai}} as the browse tool, and the same LLM as the agent model for summarization. 
More details, an example trajectory, and ablations on the search tool, browse tool, and summarization module are shown in \S\ref{app:ours}.

\section{Results}

We evaluate API models, o3, o4-mini, and Claude-4-Sonnet, open-weight models, \gptoss{} \citep{openai2025gptoss120bgptoss20bmodel} and \glm{} \citep{5team2025glm45agenticreasoningcoding}, and RL-trained models, \tongyidr{} \citep{tongyidr}.
For each instance, we evaluate the system's performance as well as the number of tool calls and tokens used.
The number of tool calls is the sum of the search API and browse/scraping operations.
Cost include the tool costs and the LLM token costs according to API prices, excluding cached tokens that are typically implemented in practical deep research systems.
For each dataset we report results averaged over all instances.
More details on the experimental setup can be found in \S\ref{app:experimental}.

\begin{figure}[t!]
    \centering
    \includegraphics[width=0.95\linewidth]{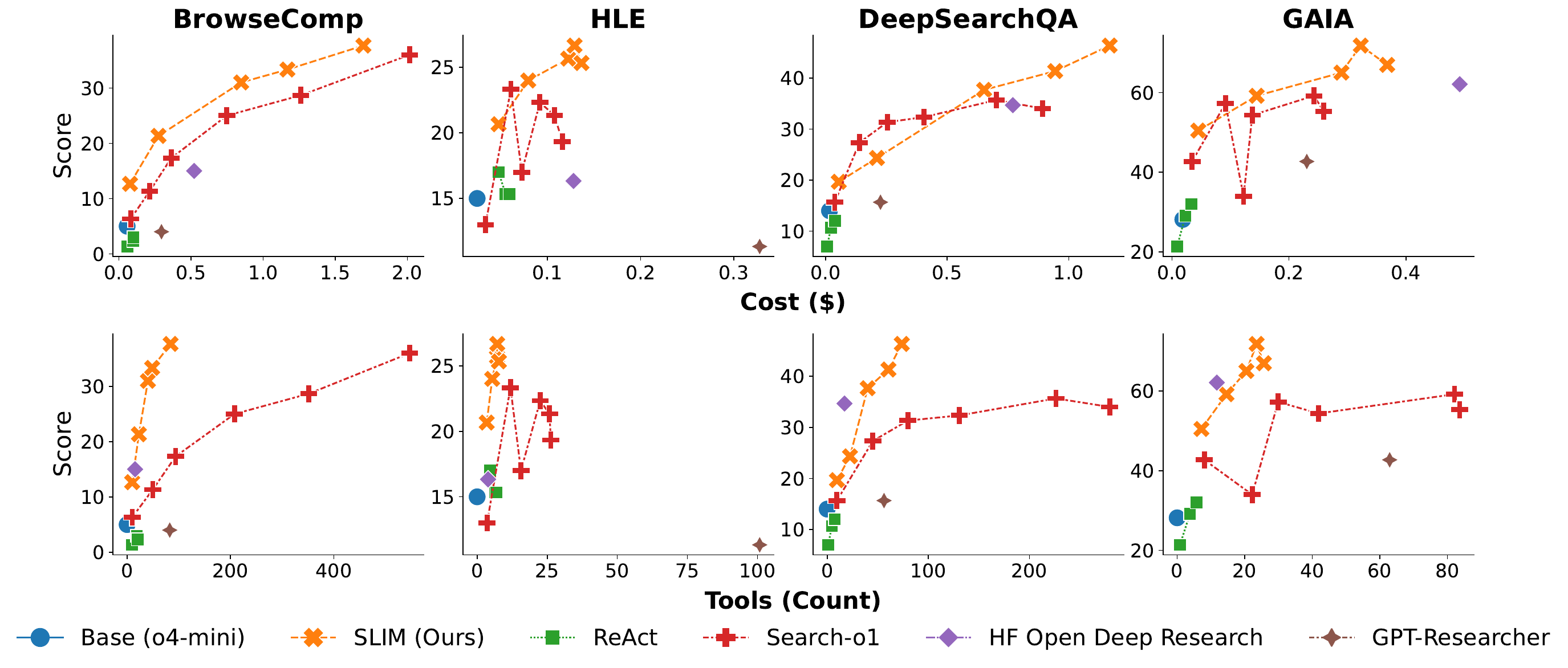}
    \vspace{-4.0pt}
    \caption{
        With o4-mini as the base model, \ours{} consistently outperforms other baselines on BrowseComp while using fewer tool calls and lower overall costs. On HLE, \ours{} can achieve overall higher performance and use fewer tool calls.
    }
    \label{fig:main_o4_mini}
\end{figure}

We present the results with o3 and o4-mini as the base models in \Cref{fig:main} and \ref{fig:main_o4_mini}.
Under the same cost with o3, \ours{} achieves significant improvements over \searcho{}, the best performing open-source framework, by 8, 6, 15, and 26 points on BrowseComp, HLE, DeepSearchQA, and GAIA, respectively.
Similarly, with o4-mini, \ours{} outperforms the open-source frameworks by 2, 4, 3, and 24 points on BrowseComp, HLE, DeepSearchQA, and GAIA, respectively.
The detailed numbers with HealthBench Hard results and comparisons are shown in \Cref{tab:main_results}.
We also conduct statistical tests to compare the performance of \ours{} with the baselines, as shown in \Cref{tab:stats_o3}.

\begin{table}[tbp]
    \centering
    \resizebox{\linewidth}{!}{%

    \begin{NiceTabular}{clc|*{15}{w{c}{0.75cm}}}
        \toprule
        & \Block{2-1}{\textbf{Model}} & \Block{2-1}{\textbf{T}} & \multicolumn{3}{c}{\textbf{BrowseComp}} & \multicolumn{3}{c}{\textbf{HLE}} & \multicolumn{3}{c}{\textbf{DeepSearchQA}} & \multicolumn{3}{c}{\textbf{GAIA}} & \multicolumn{3}{c}{\textbf{HealthBench Hard}} \\
        \cmidrule(lr){4-6}\cmidrule(lr){7-9}\cmidrule(lr){10-12}\cmidrule(lr){13-15}\cmidrule(lr){16-18}
        & & & Tools & Cost & Score & Tools & Cost & Score & Tools & Cost & Score & Tools & Cost & Score & Tools & Cost & Score \\
        \midrule\midrule
        \Block[fill=red!8]{4-1}{\rotatebox{90}{\textbf{GLM-4.7}}} & Base & - & 2.7 & 0.9 & 2.0 & 1.8 & 0.8 & 9.0 & 3.1 & 0.7 & 8.0 & 2.6 & 0.8 & 18.4 & 1.5 & 0.3 & 12.9 \\
        & \react & 10 & 27.8 & 3.0 & 2.0 & 17.8 & 2.2 & 9.0 & 30.1 & 2.9 & 8.0 & 24.7 & 2.6 & 18.4 & 15.6 & 1.5 & 12.9 \\
        & \searcho & 50 & 155.3 & 15.9 & 7.7 & 49.4 & 5.3 & 9.7 & 116.0 & 11.7 & 19.0 & 69.0 & 7.0 & 45.6 & 23.0 & 2.3 & 11.5 \\
        & \ours & 150 & 51.8 & 4.9 & \textbf{17.3} & 19.8 & 2.5 & \textbf{13.7} & 48.8 & 5.1 & \textbf{20.7} & 26.0 & 2.7 & \textbf{49.5} & 5.2 & 0.5 & \textbf{13.7} \\
        \midrule
        \Block[fill=blue!8]{4-1}{\rotatebox{90}{\textbf{GPT-OSS}}} & Base & - & 0.0 & 0.2 & 2.7 & 0.0 & 0.2 & 10.0 & 0.0 & 0.1 & 9.3 & 0.0 & 0.1 & 23.3 & 0.0 & 0.1 & 36.6 \\
        & \react & 10 & 26.1 & 3.2 & 3.3 & 10.3 & 1.4 & 11.0 & 34.8 & 4.3 & 4.7 & 25.0 & 3.1 & 30.1 & 4.8 & 0.8 & 39.1 \\
        & \searcho & 50 & 45.7 & 4.4 & 6.0 & 9.7 & 1.1 & 16.0 & 60.1 & 5.8 & 15.0 & 29.1 & 2.9 & 45.6 & 2.6 & 0.4 & 37.6 \\
        & \ours & 150 & 70.5 & 12.5 & \textbf{20.0} & 8.1 & 1.3 & \textbf{19.0} & 61.6 & 11.6 & \textbf{34.3} & 23.5 & 3.9 & \textbf{58.3} & 2.2 & 0.4 & \textbf{39.3} \\
        \bottomrule
    \end{NiceTabular}

    }
    \caption{\ours{} performs the best across all settings. We evaluate open-weight models with settings of comparable cost. Costs are in US cents. More settings are available in \S\ref{app:additional_results}.}
    \label{tab:main_open}
\end{table}

\begin{wraptable}{r}{0.48\linewidth}
    \vspace{-12pt}
    \centering
    \small
    \resizebox{\linewidth}{!}{

    \begin{tabular}{llllllll}
        \toprule
        & & \multicolumn{3}{c}{\textbf{BrowseComp}} & \multicolumn{3}{c}{\textbf{HLE}}  \\
        \cmidrule(lr){3-5} \cmidrule(lr){6-8}
        &  & Tools & Cost & Score & Tools & Cost & Score \\
        \midrule
        Tongyi-DR-30B & - & 0.0 & 2.8 & 2.3 & 0.0 & 2.2 & 11.0 \\
        \react & 10 & 39.0 & 3.5 & 1 & 28.0 & 3.0 & 12 \\
        \searcho & 10 & 70.2 & 11.1 & 14.3 & 44.4 & 8.4 & 20.0 \\
        \ours & 150 & 61.6 & 8.3 & 19.7 & 23.1 & 5.3 & 19.7\\
       \bottomrule
    \end{tabular}

    }

    \caption{
        \ours{} significantly outperforms \searcho{} on BrowseComp and achieves similar performance on HLE with lower cost.
        Costs are in US cents.
    }
    \label{tab:results_tongyi}
    \vspace{-1em}
\end{wraptable}

We also show results with different base models---\glm{} and \gptoss{} in \Cref{tab:main_open}, \tongyidr{} in \Cref{tab:results_tongyi}, and Claude-4-Sonnet in \Cref{fig:main_claude}.
\ours{} consistently achieves the better or comparable performance at lower cost across these models and all datasets compared to other frameworks, suggesting that our simple design generalizes well to models of different sizes and training strategies.
Notably, \ours{} shows consistent trends across all three base models whereas certain frameworks only work well under certain models and datasets; for instance, \hfodr{} performs well on DeepSearchQA and GAIA but poorly on BrowseComp, HealthBench, and HLE.
Overall, this is strong evidence that \ours{} serves as an effective framework for long-horizon tasks.
Full results and additional ablations on different context management strategies (e.g., discarding context) and other design choices are in \S\ref{app:additional_results}.

\section{Fine-Grained Trajectory-level Analysis}

\subsection{Trajectory-Level Error Taxonomy}
\label{sec:trajectory-level-analysis}

\begin{figure}[t!]
    \centering
    \includegraphics[width=0.9\linewidth]{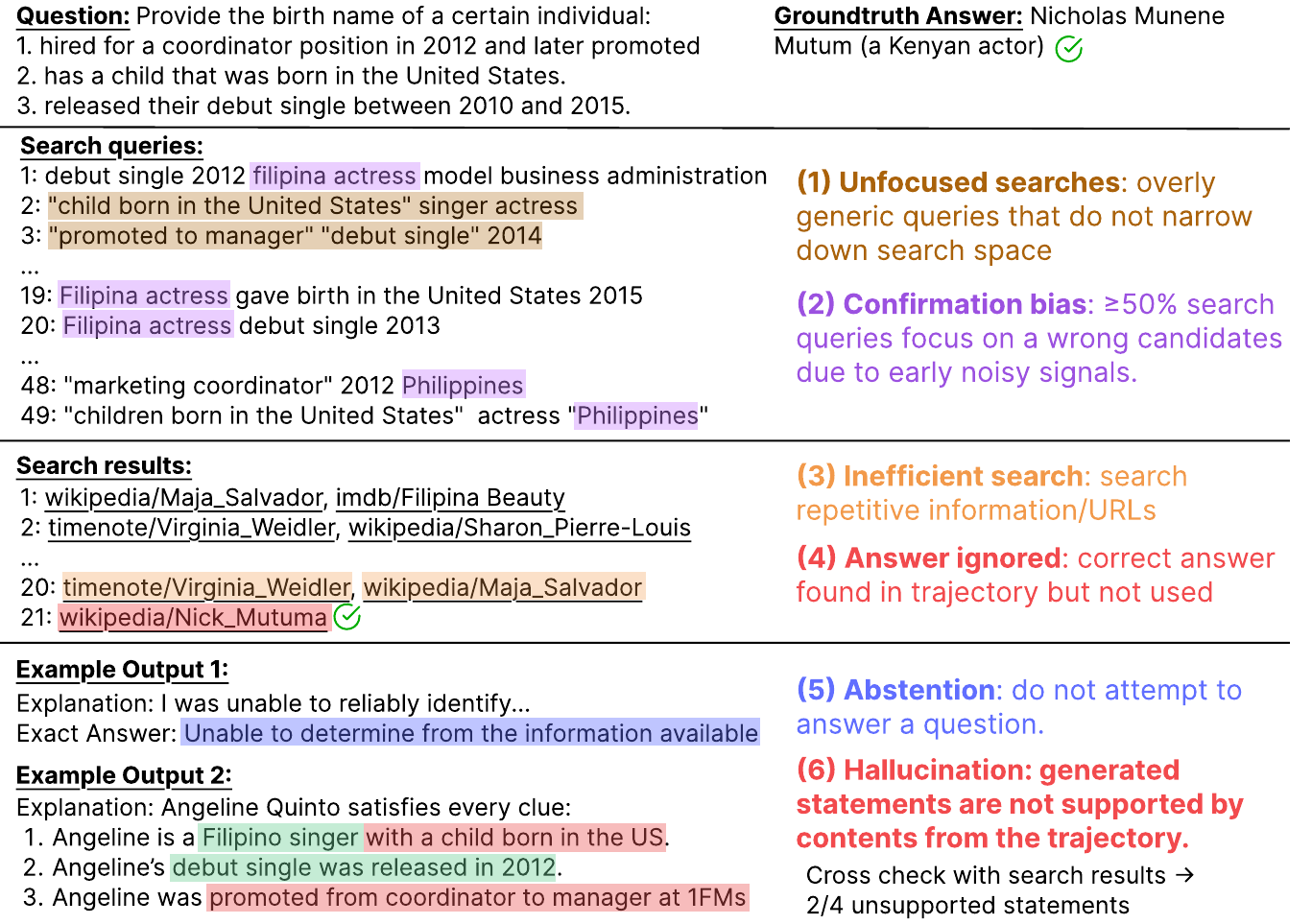}
    \vspace{-8.0pt}
    \caption{
        Examples of each trajectory-level failure mode on a BrowseComp sample. 
    }
    \label{fig:error_examples}
\end{figure}

To understand how \ours{} improves over other systems at a deeper level, we extend the analysis beyond the task outcome, and focus on characterizing the mistakes that a system makes over the course of its long search \emph{trajectories}.
To this end, we first develop a shared taxonomy of common failure modes by manually examining individual trajectories from the compared systems on BrowseComp.
We present examples of each failure mode in the taxonomy in \Cref{fig:error_examples}, and detailed definitions in \S\ref{app:error}.
Our taxonomy covers possible failure modes for long-horizon search agents in the information gathering process (e.g., unfocused searches, confirmation bias, and inefficient search) as well as the answer synthesis stage (e.g., ignoring the answer, abstention, and hallucination).

Based on the taxonomy, we develop an automated error analysis pipeline that annotates each trajectory with the failure modes using a mix of rule-based heuristics and LLM-as-a-judge approaches.
Our pipeline carefully examines all parts of each trajectory---the search queries and results, the browsed contents, and the final answer---to identify the failure modes.
We describe the pipeline more in \S\ref{app:error}.

\subsection{Analysis of Trajectory-level Failure Modes}
For fair comparison, we analyze all frameworks under a similar cost budget\footnote{We exclude \gptr{} because their implementation do not return the contents of the search results.}: we choose the setting with the closest cost to \ours{} with tool budget $T = 150$, according to \Cref{tab:main_results}.
The distribution of trajectory-level errors are shown in \Cref{tab:analysis_errors}, where we show the percentage of correct answer and each failure mode across all samples.
We first observe that \ours{}'s advantage in performance could be attributed to the notably reduced hallucination rate compared to other frameworks.
This is likely due to the fact that \ours{} can choose what URLs to browse based on the search results, allowing it to reduce the amount of noise in the context. 
In contrast, the other frameworks observe significantly higher hallucination rates compared to \ours{}, suggesting that they often resort to their parametric knowledge to answer the question when they cannot find the correct answer through tool calls.

Moreover, \searcho{} and \ours{} observe higher percentages of answer ignored than other frameworks.
One explanation is that these frameworks tend to encounter more search results across their longer trajectories, which leads to a higher chance of finding the answer, but also a higher chance of ignoring it.
In contrast, \react{} and \hfodr{} do not scale well to longer trajectories, which means they are unlikely to encounter the correct answer.
Our analysis reveals that a promising direction for improving long-horizon agentic search is to enable language models to better identify the correct answer from long trajectories.

Notably, despite the improvements on hallucination, \ours{} still suffers from high abstention rates, and is more prone to ignoring the groundtruth answers.
We leave these improvements to future work, and hope that our trajectory-level analysis can be a useful tool for improving long-horizon systems in more interpretable and concrete ways.
\begin{table}[!t]
    \vspace{-1em}
    \centering
    \small
    \resizebox{0.98\linewidth}{!}{
    \begin{tabular}{lrrrrrrrr}
        \toprule
        &\multirow{2}{4em}{\raggedleft \textbf{Turn Budget}}& \multirow{2}{4em}{\raggedleft \textbf{Correct}} & \multirow{2}{4em}{\raggedleft \textbf{Confirm Bias}} & \multirow{2}{4em}{\raggedleft \textbf{Unfocused Search}} & \multirow{2}{4em}{\raggedleft \textbf{Inefficient Search}} & \multirow{2}{4em}{\raggedleft \textbf{Abstention}} & \multirow{2}{4em}{\raggedleft \textbf{Answer Ignored}} & \multirow{2}{4em}{\raggedleft \textbf{Hallucinate}} \\
        \textbf{Framework} & & & & & & & \\
        \midrule

        \react & 10 & 7.0 & 9.3 & 44.0 & 3.9 & 1.0 & 0.7 & 56.7 \\
        \searcho & 50 & 48.3 & 9.3 & 33.7 & 7.2 & 4.3 & 26.0 & 46.8 \\
        \hfodr & 20 & 20.0 & 6.7 & 58.7 & 43.9 & 32.3 & 1.7 & 96.2 \\
        \ours & 150 & 56.0 & 9.7 & 34.0 & 7.6 & 27.7 & 30.7 & 19.0 \\

        \bottomrule
    \end{tabular}
    }
    \caption{
        The percentage of trajectory over all samples that observe each failure mode.
        For hallucination only, we report the percentage of hallucinations for samples that ends with an incorrect answer and do not abstain.
    }
    \label{tab:analysis_errors}

\end{table}

\section{Related Work}

\paragraph{Long-horizon search agents.}
Recently, the community has taken great interests in search agents that extend beyond single-turn RAG by interleaving web search with reasoning across multiple turns~\citep{yao2023react}. This paradigm has been actively developed in both industry \citep{OpenAIDeepResearch2025,GoogleGeminiDeepResearch2025,xAI_Grok32025,nguyen2025sfrdeepresearcheffectivereinforcementlearning} and open-source communities \citep[][\ti{inter alia}]{jin2025searchr,li2025searcho1agenticsearchenhancedlarge,li2025websailornavigatingsuperhumanreasoning, tongyidr}, advancing agent capabilities. However, as trajectories grow longer, these agents accumulate noisy context and become bounded by the context window, forcing early termination before tasks are fully resolved~\citep{liu2024lost,hong2025context}. To address this, recent efforts explore context management strategies that compress or prune trajectory history, preserving capacity for future steps. 
Concurrent to us, several works explore strategies such as \textit{Summary}~\citep{wu2025resumunlockinglonghorizonsearch}, \textit{Discard-All}~\citep{liu2025deepseek}, and \textit{Keep-Recent-N}~\citep{tang2025beyond,zeng2026glm}, yet no clear winner has been observed. 
In contrast, we take a more holistic approach of long-horizon agentic search framework by carefully designing both the tool interfaces and context management. 
Notably, we also provide rigorous error analysis pipeline, enabling systematic comparison across different methodologies.

\paragraph{RL for long-horizon systems.}
Many recent works seek to improve search agents through reinforcement learning \citep[][\ti{inter alia}]{Li2025WebThinker,zheng2025deepresearcherscalingdeepresearch,chen2025researchlearningreasonsearch,li2025websailornavigatingsuperhumanreasoning,wu2025resumunlockinglonghorizonsearch}.
A popular approach is to synthetically generate question-answer pairs that require long-horizon search trajectories \citep{xia2025opendatasynthesisdeep,tao2025webshaperagenticallydatasynthesizing}.
Other works focus on comparing different training objectives \citep{jin2025searchr,jin2025empiricalstudyreinforcementlearning}.
However, critical analysis of the error modes and comparison of different frameworks are still lacking.

\section{Conclusion}

In this work, we propose \ours, a simple yet effective long-horizon agentic search framework that addresses context limitations prevalent in existing systems.
We show that \ours{} consistently achieves the highest performance across different proprietary and open models and diverse datasets compared to other frameworks while using fewer tool calls and lower overall costs, suggesting that our simple design enables better long-horizon agentic search. 
We also develop an automated error analysis pipeline to characterize the failure modes of long-horizon systems.
Our analysis shows that \ours{} is more resistant to failure modes such as hallucination.
We hope our framework and analysis pipeline can serve as a useful tool for the community to understand and improve long-horizon agentic search systems.

\bibliography{custom}
\bibliographystyle{colm2026_conference}

\appendix
\section{Appendix}
\label{sec:appendix}

\subsection{Existing Frameworks}
\label{app:existing}

\paragraph{\react} \citep{yao2023react} is a simple framework that allows an LLM agent to alternate between thinking and acting.
This framework allows the agent to use tool calls across many turns.
Following the original work's knowledge-intensive task settings, our implementation gives the LLM access to a single search tool---given a query, the tool returns a list of top 10 search results, from a search engine, along with their web contents.
The search results are then concatenated and appended to the agent context for subsequent steps.
When the LLM chooses not to use the search tool, the final output is used for evaluation.

In our implementation, we vary the maximum number of turns in each trajectory from 1 to 10.
Consistent with \ours, we use Google as the search engine, accessed through the Serper API, which returns a list of top 10 search results.
Each search result contains a title, a URL, and a short snippet of the content.
After obtaining the top 10 search results, we emulate previous RAG approaches by scraping all search result URLs and concatenate their content.
Similar to \ours, we use crawl4ai to scrape web pages.
We truncate each scraped document to at most 10,000 characters, which corresponds to roughly 1,000 tokens.

We notice that \react{} often hits the context window limit as the retrieval results are often too long.
When the LLM API call fails due to the context window limit, we fallback to not using any tools and just ask the base LLM to answer the question.
As a result, we only experiment with up to 10 turns, where the framework already falls back to not using any tools for most queries.
A sketch of the framework is shown in Alg. \ref{alg:react}.

\paragraph{\searcho} \citep{li2025searcho1agenticsearchenhancedlarge} builds upon \react{} with an additional ``reason-in-document'' step, where an LLM summarizes the list of top 10 search results and their contents before appending the results to the agent's input context.
Although the summary added to the agent context is relatively short compared to the full search result, this approach still uses a large amount of browse calls in each search step, and the summarization steps incur a large amount of LLM token usage.
In our setting, we vary the maximum number of turns in each trajectory from 1 to up to 100 turns.

Similar to \react, the retrieval tool at each step consists of a single Serper API call, followed by multiple scraping operations.
We adopt the code from the original implementation\footnote{\url{https://github.com/RUC-NLPIR/Search-o1}}, which uses BeautifulSoup\footnote{\url{https://beautiful-soup-4.readthedocs.io/en/latest/}} to scrape the search result URLs.
In this implementation, the scraping operation will extract part of the web content that best matches the short snippet returned by the search engine.
The matching is done by simply computing the F1 scores between the snippet and sentences in the web page.
Subsequently, the context is filled up with at most 2,500 characters from the web page.
Then, all context from the search results are concatenated and appended to the agent context for the summarization step.

It is important to note that the scraping operation is relatively expensive due to the network latency, resulting in long running time for the framework.
A sketch of the framework is shown in Alg. \ref{alg:searcho1}.

\paragraph{\hfodrfull} \citep[\hfodr;][]{huggingface2025opendeepresearch} leverages a hierarchical structure consisting of a manager agent and a search agent. 
The manager agent calls the search agent to perform detailed searches, and the search agent iteratively interacts with the search engine and a simulated browser to gather information.
When the search agent concludes its searches, it generates a summary of its searches and returns it to the manager agent.
The manager agent may use the summary to issue additional queries or output the final answer.
Furthermore, another key feature of \hfodr{} is its access to additional tools, such as a Python interpreter.
We use the default settings\footnote{\url{https://github.com/huggingface/smolagents/tree/main/examples/open_deep_research}}, which fixes the maximum number of turns for the manager and search agent to be 20.
A sketch of the framework is shown in Alg. \ref{alg:hfodr}.
Specific descriptions of each tool can be found in \Cref{tools:hfodr}.

\paragraph{\gptrfull} \citep[\gptr;][]{Elovic_gpt-researcher_2023} is a complex multi-agent system where each agent has distinct roles.
Specifically, the system consists of a researcher conductor that orchestrates the search process, a report generator that generates the final report at the end of the search process, a context manager that summarizes search results, and a source curator that selects relevant sources from scraped web pages.
Finally, \gptr{} uses a deep researcher agent that acts as the node of a search tree, where each node is able to spawn multiple child nodes, each of which is a system with the previously described components.
We use the default settings of the framework\footnote{\url{https://github.com/assafelovic/gpt-researcher}}, which fixes the depth of the search tree to be 2 and the breadth of search at each depth to be 4.
A sketch of the framework is shown in Alg. \ref{alg:gptr}.

\paragraph{Other frameworks.}
There are many recent works on agentic search systems and memory-management frameworks \citep{gangi-reddy-etal-2025-infogent,xu2025comprehensivesurveydeepresearch,belcak2025universaldeepresearchbring}. We chose the most popular open-source agentic search and deep research systems for comparison. These systems also span both simple single-agent and complex multi-agent systems, which we believe serve as a representative and fair group of baselines for the paper. Due to the high cost and long runtime of agentic systems, we only evaluate the representative baselines. Although there are explicit memory-management frameworks, we find that existing summarization models already do something similar to memory-selective mechanisms through qualitative analysis. In the example trajectory we show in \Cref{fig:slim-detailed}, the model summarizes the trajectory into several bullet points, such as ``Investigation and findings so far'', ``Current hypothesis'', and ``Needed next''. The resulting summary is similar to many memory-selective mechanisms that only retain relevant facts to the current query. Thus, we find that allowing the model to compress the full trajectory naturally filters out irrelevant information while achieving simplicity and avoiding over-prompt-engineering.

\SetKwComment{Comment}{/* }{ */}
\SetKwProg{Fn}{Function}{:}{}

\begin{algorithm}[hbt!]
\caption{ReAct}\label{alg:react}
\KwData{Task input $x$, LLM $\theta$, maximum number of turns $T$}

\Fn{\text{search}($q$)}{
    \Return{(title$_i$, url$_i$, snippet$_i$)}$_{i=1}^k$\;
}
\Fn{\text{browse}($u$, $q$)}{
    $D \gets \text{scrape}(u)$\;
    \Return{$D[:10000]$}\;
}

\KwResult{Task output $y$}
Turn $t \gets 1$\;
Context $C \gets \{x\}$\;
$\mathcal{T} \gets \{\text{search}\}$\;

\While{$t < T$}{
    $o_t \gets \theta(C; \mathcal{T})$ \Comment*[r]{LLM may only call the search tool}
    \Switch{$o_t$}{
        \Case{\text{search}}{
            $R \gets \text{search}(o_t)$ \Comment*[r]{Perform search}
            $C \gets C \cup \{o_t\}$\Comment*[r]{Browse every search result and append}
            \For{$(t_i, u_i, s_i) \in R$}{
                $C \gets C \cup \text{browse}(u_i, s_i)$ 
            }
        }
        \Case{Final Answer}{
            \Return{$o_t$}\;
        }
    }
    $t \gets t + 1$\;
}
\Return{$\theta(C; \text{final answer})$}\;
\end{algorithm}

\SetKwComment{Comment}{/* }{ */}
\SetKwProg{Fn}{Function}{:}{}

\begin{algorithm}[hbt!]
\caption{Search-o1}\label{alg:searcho1}
\KwData{Task input $x$, LLM $\theta$, maximum number of turns $T$, summary interval $n$}

\Fn{\text{search}($q$)}{
    \Return{(title$_i$, url$_i$, snippet$_i$)}$_{i=1}^k$\;
}
\Fn{\text{visit}($u$, $q$)}{
    $D \gets \text{scrape}(u)$\;
    $D \gets \text{split}(D) = \{d_i\}_{i=1}^m$\;
    if $q = \emptyset$ then
        \Return{$d' \gets d_1$}\;
    else
        $d' \gets \argmax_{d_i \in D} \text{F1}(d_i, q)$\;
    \Return{$d'$}\;
}

\KwResult{Task output $y$}
Turn $t \gets 1$\;
Context $C \gets \{x\}$\;
$\mathcal{T} \gets \{\text{search}\}$\;

\While{$t < T$}{
    $o_t \gets \theta(C; \mathcal{T})$ \Comment*[r]{LLM may only call the search tool}
    \Switch{$o_t$}{
        \Case{\text{search}}{
            $R \gets \text{search}(o_t)$ \Comment*[r]{Perform search}
            $l \gets \text{length}(C)$\;
            $D \gets \{c_i\}_{i=l-5}^l$ \;
            \For{$(t_i, u_i, s_i) \in R$}{
                $D \gets D \cup \text{visit}(u_i, s_i)$ \Comment*[r]{Visit every search result}
            }
            $C \gets C \cup \{o_t, \theta(D; \text{summarize})\}$\;
        }
        \Case{Final Answer}{
            \Return{$o_t$}\;
        }
    }
    $t \gets t + 1$\;
}
\Return{$\theta(C; \text{final answer})$}\;
\end{algorithm}

\SetKwComment{Comment}{/* }{ */}
\SetKwProg{Fn}{Function}{:}{}

\begin{algorithm}[hbt!]
\caption{HuggingFace Open Deep Research}\label{alg:hfodr}
\KwData{Task input $x$, LLM $\theta$, maximum number of turns for search and main agents $T_s$ and $T_m$, respectively, and planning interval $p$}
$\text{web\_tools}\gets \{\text{Search},\text{Visit},\text{Page Up},\text{Page Down},\text{Finder},\text{Find Next},\text{Archive Search},\text{Text Inspector}\}$\;
$\text{main\_tools}\gets \{\text{search\_agent},\text{Visualize},\text{Text Inspector}\}$\;

\Fn{\text{plan}($q$, $c$)}{
    \Comment{Prompt the LLM to generate a plan}
    \Return{$\theta(q, c; \text{plan})$}\;
}

\Fn{\text{search\_agent}($q$)}{
    $P \gets \text{plan}(q, \emptyset)$\;
    $C \gets \{q, P\}$\;
    $t \gets 1$\;
    \While{$t < T_s$}{
        \If{$t \mod p = 0$}{
            $P \gets \text{plan}(q, C)$\;
            $C \gets C \cup \{P\}$\;
        }
        $o_t \gets \theta(C; \text{web\_tools})$\;
        \If{$type(o_t) = \text{final\_answer}$}{
            \Return{$o_t$}\;
        }
        \Comment{do the tool call, see \ref{tools:hfodr} for tool details}
        $C \gets C \cup \{o_t, \text{tool}(o_t)\}$\;
        $t \gets t + 1$\;
    }
    \Return{$\theta(C; \text{final answer})$}\;
}

\KwResult{Task output $y$}
Turn $t \gets 1$\;
$P \gets \text{plan}(x, \emptyset)$\;
Context $C \gets \{x, P\}$\;

\Comment{the main agent plans and calls the search agent}
\While{$t < T_m$}{
    \If{$t \mod p = 0$}{
        $P \gets \text{plan}(x, C)$\;
        $C \gets C \cup \{P\}$\;
    }
    $o_t \gets \theta(C; \text{main\_tools})$\;
    \If{$type(o_t) = \text{final\_answer}$}{
        \Return{$o_t$}\;
    }
    $C \gets C \cup \{o_t, \text{tool}(o_t) \}$\;
    $t \gets t + 1$\;
}
\Return{$\theta(C; \text{final answer})$}\;
\end{algorithm}

\SetKwComment{Comment}{/* }{ */}
\SetKwProg{Fn}{Function}{:}{}

\begin{algorithm}[hbt!]
\caption{GPT-Researcher}\label{alg:gptr}
\KwData{Task input $x$, LLM $\theta$, research depth $D$, research breadth $B$, summary interval $n$}

\Fn{\text{search}($q$)}{
    \Return{(title$_i$, url$_i$, snippet$_i$)}$_{i=1}^k$\;
}
\Fn{\text{visit}($u$, $q$)}{
    $D \gets \text{scrape}(u)$\;
    $D \gets \text{split}(D) = \{d_i\}_{i=1}^m$\;
    if $q = \emptyset$ then
        \Return{$d' \gets d_1$}\;
    else
        $d' \gets \argmax_{d_i \in D} \text{F1}(d_i, q)$\;
    \Return{$d'$}\;
}
\Fn{\text{plan}($q$)}{
    \Comment{Prompt the LLM to generate a list of queries}
    $R \gets \text{search}(q)$\;
    \Return{$\theta(x, R; \text{plan})$}\;
}
\Fn{\text{conduct\_research}($q$)}{
    \Comment{Conduct research on one query by generating subqueries and retrieve and scrape}
    $Q \gets \text{plan}(q)$\;
    $R \gets \emptyset$\;
    \For{$q_i \in Q$}{
        \For{$t_i, u_i, s_i \in \text{search}(q_i)$}{
            $r_i \gets \text{visit}(u_i, s_i)$\;
            $R \gets R \cup r_i$\;
        }
    }
    \Return{$\theta(x, R; \text{process})$}\;
}
\Fn{\text{deep\_research}($q$, $d$)}{
    \Comment{Recursively plan and conduct research}
    $Q \gets \text{plan}(q)$\;
    $R \gets \emptyset$\;
    \For{$q_i \in Q$}{
        $r_i \gets \text{conduct\_research}(q_i)$\;
        \Comment{Prompt the LLM to generate takeaways and follow up questions}
        $q_i' \gets \theta(r_i; \text{process})$\;
        \If{$d < D$}{
            $R \gets R \cup \text{deep\_research}(q_i', d+1)$\;
        }
    }
    \Return{$R$}\;
}

\KwResult{Task output $y$}
Turn $t \gets 1$\;
Context $C \gets \{x\}$\;

$P \gets \text{plan}(x)$\;
$R \gets \text{deep\_research}(P, 1)$\;

\Return{$\theta(R; \text{write report})$}\;
\end{algorithm}

\subsection{HuggingFace Open Deep Research Tools}
\label{tools:hfodr}

\hfodr{} is a hierarchical framework that consists of a manager agent and a search agent.
The manager agent has access to the following tools:
\begin{enumerate}
    \item \textbf{Search Agent}: an agent that will search the internet to answer a question.
    \item \textbf{Visualizer}: given the path to a downloaded image, it will call an LLM to answer questions about the image.
    \item \textbf{Text Inspector}: given the path to a downloaded text file, it will call an LLM to answer questions about the text.
\end{enumerate}

The search agent has access to the following tools:
\begin{enumerate}
    \item \textbf{Google Search}: a search engine that will search the internet to answer a question. This tool uses Serper API in the backend.
    \item \textbf{Visit Tool}: visit a URL and render the page in HTML as in a browser.
    \item \textbf{Page Up}: navigate the current page by scrolling up.
    \item \textbf{Page Down}: navigate the current page by scrolling down.
    \item \textbf{Finder Tool}: find a text in the current page.
    \item \textbf{Find Next}: find the next occurrence of the text in the current page.
    \item \textbf{Archive Search}: search the archives for information. 
    \item \textbf{Text Inspector}: given the path to a downloaded text file, it will call an LLM to answer questions about the text.
\end{enumerate}

Detailed descriptions of each tool can be found in the original implementation\footnote{\url{https://github.com/huggingface/smolagents/blob/main/src/smolagents/default_tools.py}}.

\subsection{Trajectory-Level Analysis Definitions}
\label{app:error}

In this subsection, we describe how we annotate each trajectory with the failure modes.
For LLM-as-a-judge approaches, we use \texttt{o3-2025-04-16} as the judge model.
In each of the following LLM-as-a-judge approaches, we use the same judge model, and force the model to generate its response in a json format for easy parsing.
We find that existing frontier LLMs are powerful enough to reliably check for simple yes/no questions and output them in a json format.

\paragraph{Confirmation bias.} 
Confirmation bias occurs when the system finds a potential candidate that is incorrect in its search process, and subsequently spends the majority of its search budget on the same candidate without considering other options, leading to a lack of exploration in the search space.
To detect this, we first collect all the search queries that the system has made and then use an LLM to check if the search queries overly focus on a single wrong candidate.
The judge model is given access to the groundtruth answer and the search queries, so it's able to determine if the search queries are relevant to the groundtruth answer and the similarities between different search queries.
We consider a trajectory to have confirmation bias if a majority of the search queries are similar to each other, and focuses on a single wrong candidate.
The prompt used for confirmation bias detection is shown in \Cref{prompt:confirmation-bias}.

\begin{table}[h]
    \centering
    \begin{tabular}{|p{0.9\textwidth}|}
    \hline
    \textbf{Prompt for Confirmation Bias Detection} \\
    \hline
    You are a helpful assistant that can analyze the trajectory of an information-seeking agent. You are given a question-answer pair and the search history of an agent that tried to answer the question. You should analyze the search history and determine if the agent spends more than half of the tool calls searching for the same incorrect answer. That is, the agent continues searching for the same topic even though it's not the correct answer to the question, and spends half or more of its tool calls on these searches. Output your final conclusion with your reasoning and a single word: 'yes' if the agent spends more than half of its tool calls on the same incorrect answer or 'no' if the agent does not.

    \textbf{Reasoning:} explain what the agent did, and if it did or did not focus its searches on a wrong answer. \\[0.5em]
    \textbf{Conclusion:} ``yes'' or ``no''. \\
    \textbf{Search queries:} \texttt{<search-queries>} \\
    \textbf{Question:} \texttt{<question>} \\
    \textbf{Correct Answer:} \texttt{<correct-answer>} \\
    \hline
    \end{tabular}
    \caption{System prompt used for detecting confirmation bias in agent trajectories}
    \label{prompt:confirmation-bias}
\end{table}

\begin{table}[h]
    \centering
    \begin{tabular}{|p{0.9\textwidth}|}
    \hline
    \textbf{Prompt for Unfocused Search Detection} \\
    \hline
    You are a helpful assistant that can analyze the trajectory of an information-seeking agent. You are given a question-answer pair and the search history of an agent that tried to answer the question. You should analyze the search history and determine if the search queries do not help the agent narrow down the search space. Consider the following cases:\\
    1. The agent searches for information relevant to the question and answer, but it's not specific enough to yield helpful results. \\
    2. The agent searches for queries that are not sufficiently relevant or specific to the question and answer, which does not narrow down the search space enough.\\
    3. The agent explores the search space with diverse queries but does not use enough tool calls to properly narrow down the search space by either eliminating wrong answers or verifying the correct answer.\\
    All of these cases are considered to be unfocused search. You should consider the whole trajectory of the agent, and not just some of the tool calls---only consider the trajectory to be unfocused if more than half of the searches are unfocused.\\

    Output your final conclusion with your reasoning and a single word: 'yes' if the searches are unfocused or 'no' if the searches are focused enough.

    \textbf{Reasoning:} explain what the agent did, and if it did or did not use tool calls to properly narrow down the search space. \\[0.5em]
    \textbf{Conclusion:} ``yes'' or ``no''. \\
    \textbf{Search queries:} \texttt{<search-queries>} \\
    \textbf{Question:} \texttt{<question>} \\
    \textbf{Correct Answer:} \texttt{<correct-answer>} \\
    \hline
    \end{tabular}
    \caption{System prompt used for detecting unfocused search in agent trajectories}
    \label{prompt:unfocused-search}
\end{table}

\begin{table}[h]
    \centering
    \begin{tabular}{|p{0.9\textwidth}|}
    \hline
    \textbf{Prompt for Groundtruth Ignored Detection} \\
    \hline
    You are a helpful assistant that can analyze the trajectory of an information-seeking agent. You are given a question-answer pair and a list of webpages. You should analyze the web contents and determine if it contains the correct answer. The correct answer is considered to be found if there are some context in the search results that is either a direct or near-exact match to the correct answer. Output your final conclusion with your reasoning and a single word: 'yes' if the content contains the correct answer or 'no' if the content does not contain the correct answer.

    \textbf{Reasoning:} explain if the web content contains the correct answer. \\[0.5em]
    \textbf{Conclusion:} "yes" or "no".\\

    \texttt{<tool-responses>} \\
    \textbf{Question:} \texttt{<question>} \\
    \textbf{Correct Answer:} \texttt{<correct-answer>} \\
    \hline
    \end{tabular}
    \caption{System prompt used for detecting groundtruth ignored in agent trajectories}
    \label{prompt:groundtruth-ignored}
\end{table}

\begin{table}[h]
    \centering
    \begin{tabular}{|p{0.9\textwidth}|}
    \hline
    \textbf{Prompt for Giving Up Detection} \\
    \hline
    You are a helpful assistant that can analyze the final output of an information-seeking agent. You are to check if the agent decides that it cannot find the correct answer. For example, if the explanation states that it cannot find enough relevant information to answer the question, or if the response is simply empty or "I don't know", then the agent did not attempt to answer the question. Output your final conclusion with a single word "yes" if the agent decides it did not find enough information to answer the question or "no" otherwise.

    \textbf{Conclusion:} "yes" or "no". \\

    \textbf{Final output:} \texttt{<final-output>} \\
    \hline
    \end{tabular}
    \caption{System prompt used for detecting giving up in agent trajectories}
    \label{prompt:giving-up}
\end{table}

\begin{table}[h]
    \centering
    \begin{tabular}{|p{0.9\textwidth}|}
    \hline
    \textbf{Prompt for Decomposing Explanation into Atomic Claims} \\
    \hline
    Read the given explanation and generate a list of atomic claims that are supported by the explanation. Atomic claims that are basic facts that cannot be further broken down. Generate at most 10 claims for the explanation.

Use the following as an example:
\begin{Verbatim}
Explanation:  Searching UFCStats for featherweight bouts 
where the loser landed 14 of 83 significant strikes (16.87 %
and went 0-for-4 on takedowns returns the fight Myles Jury 
vs. Ricky Glenn at UFC 219: Cyborg vs Holm (30 Dec 2017).
• Ricky Glenn (nickname "The Gladiator"—a synonym 
for swordsman) 
was the loser: sig. strikes 14/83 (16.87 %
• Both fighters (Jury 29, Glenn 28) were under 35 and 
are American.
• The referee was John McCarthy, whose first event for 
the UFC was in 1994.
Thus, the MMA event is UFC 219: Cyborg vs Holm.

Exact Answer: UFC 219: Cyborg vs Holm

Confidence: 75%
\end{Verbatim}

\textbf{Atomic Claims}: \\
- Ricky Glenn was the loser\\
- Ricky Glenn was nicknamed "The Gladiator"\\
- The sig. strike rate of Ricky Glenn was 14/83 (16.87%
- The takedown rate of Ricky Glenn was 0/4\\
- Jury was age 29\\
- Glenn was age 28\\
- Jury is American\\
- Glenn is American\\
- The referee was John McCarthy\\
- John McCarthy's first event for the UFC was in 1994\\

Output the atomic claims in the form of a json list.\\
    \hline
    \end{tabular}
    \caption{System prompt used for decomposing the model's explanation into a set of atomic claims}
    \label{prompt:decompose-explanation}
\end{table}

\begin{table}[h]
    \centering
    \begin{tabular}{|p{0.9\textwidth}|}
    \hline
    \textbf{Prompt for Hallucination Detection} \\
    \hline
    You are a helpful assistant that can analyze the trajectory of an information-seeking agent. You are given a list of webpages and a list of claims made by the agent. You should analyze the web contents to determine if each claim is supported by the web content.
A claim is supported by the web content if its factual information is mostly supported by the web content, and is not contradicted by the web content. Output your final conclusion with a list of claims that are supported by the web content. Output the list in the form of a json list, and you only need to write the index of the supported claims in the list and nothing else.

\textbf{Webpages:} \texttt{<webpages>} \\
\textbf{Atomic Claims:} \texttt{<atomic-claims>} \\
    \hline
    \end{tabular}
    \caption{System prompt used for detecting hallucination in agent trajectories}
    \label{prompt:hallucination-detection}
\end{table}

\paragraph{Unfocused search.}
Unfocused search occurs when the system generates overly generic search queries that are not useful for narrowing down the search space---the system cannot make any progress towards finding useful information.
To detect this, we first collect all the search queries that the system has made and then use an LLM to check if the search queries are generic and not useful for narrowing down the search space.
We consider a trajectory to have unfocused search if a majority of the search queries are overly generic and not useful for narrowing down the search space.
The prompt used for unfocused search detection is shown in \Cref{prompt:unfocused-search}.

\paragraph{Inefficient tool usage.}
Inefficient tool usage occurs when the system does not discover new information with its tool calls, and is therefore wasting its tool budget.
Specifically, we use URLs as a proxy for the information discovered by the system---a tool call that only return URLs seen in previous search results is considered as a waste of tool budget.
We use a simple heuristic for this analysis---iterate over all search calls made in the trajectory and keep track of seen URLs.
Then, we report the percentage of search calls that only return URLs seen in previous search results.

\paragraph{Answer ignored.}
Answer ignored occurs when the system encounters the correct answer in its search process, but does not use it to answer the question.
One possible explanation is that the system is distracted by other noisy information in its context, preventing it from correctly identifying the groundtruth.
We employ a simple approach for this analysis---we check if the groundtruth answer is present in any of the tool responses.
We employ a LLM judge to enable fuzzy matching between the groundtruth answer and the tool responses.
The prompt used for answer ignored detection is shown in \Cref{prompt:groundtruth-ignored}.
We iterate over all tool calls and use this check to determine if any tool responses contain the groundtruth answer.
We terminate the iteration if we find a tool response that contains the groundtruth answer, and report the percentage trajectories where at least one tool response contains the groundtruth answer.

\paragraph{Abstention.}
Abstention occurs when the system does not attempt to answer the question due to the lack of information in its context.
Existing LLMs can often refuse to answer the question if it is not confident in answering the question, but this behavior is not desirable for search agents that could leverage additional tool calls to find the necessary information.
We use a simple LLM judge to check if the system attempted to answer the question.
The prompt used for giving up detection is shown in \Cref{prompt:giving-up}.

\paragraph{Hallucination.}
Hallucination occurs when the system generates information that is not supported by the information it has discovered in its search process. 
In agentic search systems, it is not desirable to hallucinate information, as it could result in incorrect and misleading answers and thus affect the trustworthiness of the system.
Inspired by previous works\citep{rashkin2021measuring,bohnet2022attributed,gao2023enabling}, we check if the system hallucinates information by first decomposing the model's explanation into a set of atomic claims.
Then, we iterate through all the tool responses from the search process and check if the tool responses support all the claims.
As long as one tool response support a claim, we consider the system to not have hallucinated that claim.
In the end, we report the average percentage of unsupported claims across trajectories.
The prompt used for decomposing the model's explanation into a set of atomic claims is shown in \Cref{prompt:decompose-explanation}, and the prompt used for hallucination detection is shown in \Cref{prompt:hallucination-detection}.
These prompts are derived from previous works that show LLMs can reliably decompose texts into a set of atomic claims and check if claims are supported by a piece of text---they also achieve high agreement with human judges \citep{gao2023enabling,kamoi2023wice,yen2025helmet}.

\subsection{\ours{} Details and Ablations}
\label{app:ours}

We show an example of a \ours{} trajectory in \Cref{fig:slim-detailed}.
A sketch of the framework is also shown in Alg. \ref{alg:ours}.
Furthermore, we ablate our design choices along the following dimensions:

\begin{figure}[t!]
    \centering
    \includegraphics[width=0.98\linewidth]{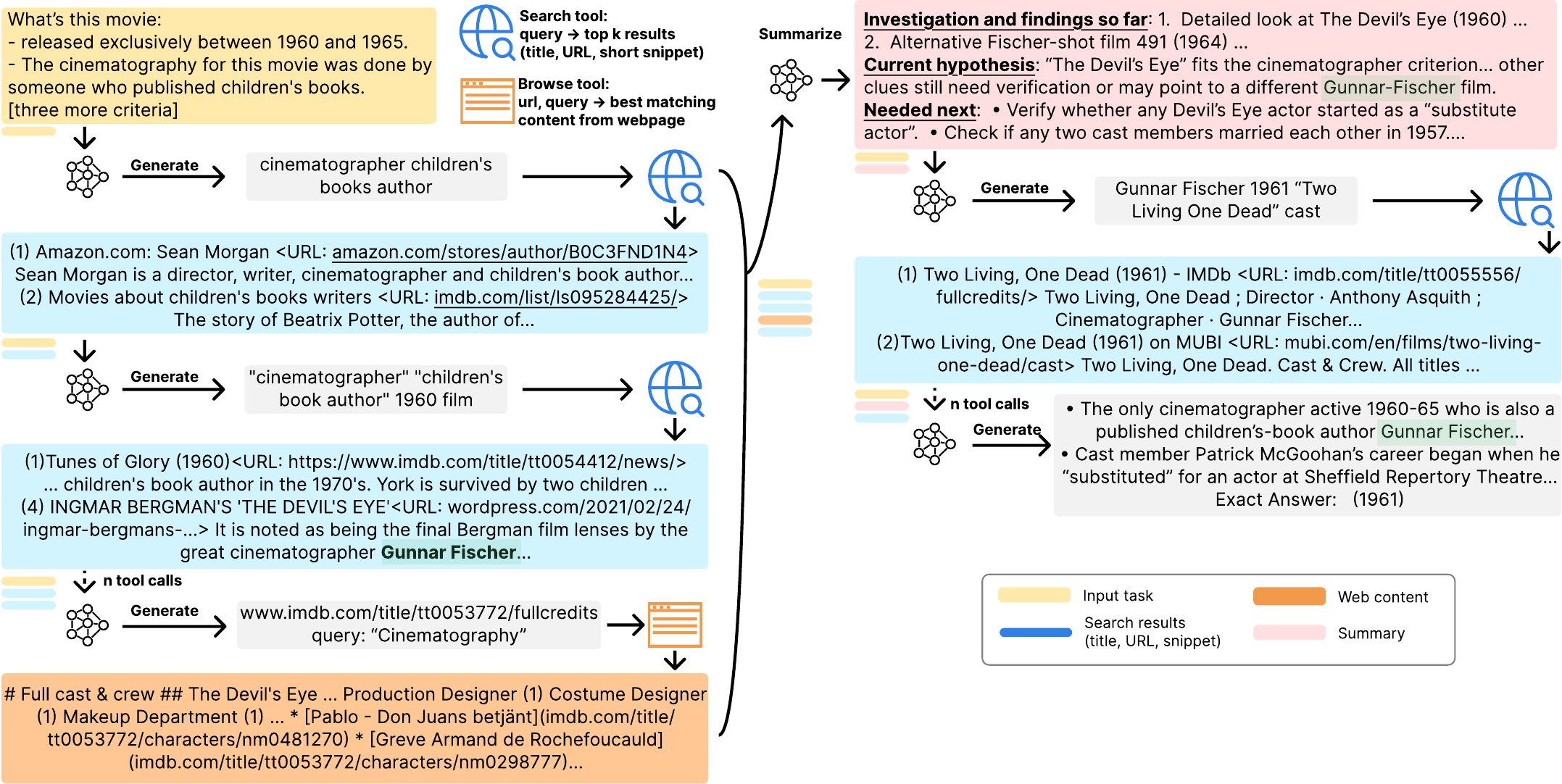}
    \vspace{-2.0pt}
    \caption{
        An example of a \ours{} trajectory.
    }
    \label{fig:slim-detailed}
\end{figure}

\SetKwComment{Comment}{/* }{ */}
\SetKwProg{Fn}{Function}{:}{}

\begin{algorithm}[hbt!]
\caption{\ours}\label{alg:ours}
\KwData{Task input $x$, LLM $\theta$, maximum number of turns $T$, summary interval $n$}

\Fn{\text{search}($q$)}{
    \Return{(title$_i$, url$_i$, snippet$_i$)}$_{i=1}^k$\;
}
\Fn{\text{browse}($u$, $q$)}{
    $D \gets \text{scrape}(u)$\;
    $D \gets \text{split}(D) = \{d_i\}_{i=1}^m$\;
    if $q = \emptyset$ then
        \Return{$d' \gets d_1$}\;
    else
        $d' \gets \argmax_{d_i \in D} \text{ROUGE-L}(d_i, q)$\;
    \Return{$d'$}\;
}

\KwResult{Task output $y$}
Turn $t \gets 1$\;
Context $C \gets \{x\}$\;
$\mathcal{T} \gets \{\text{search}, \text{browse}\}$\;

\While{$t < T$}{
    \If{$t \mod n = 0$}{ 
        $C \gets \theta(C; \text{summarize})$ \Comment*[r]{Summarize every $n$ turns}
    }

    $o_t \gets \theta(C; \mathcal{T})$\;
    \Switch{$o_t$}{
        \Case{\text{search}}{
            $q_t \gets o_t$\; 
            $C \gets C \cup \{o_t, \text{search}(q_t)\}$\;
        }
        \Case{\text{browse}}{
            $u_t, s_t \gets o_t$\;
            $C \gets C \cup \{o_t, \text{browse}(u_t, s_t)\}$\;
        }
        \Case{Final Answer}{
            \Return{$o_t$}\;
        }
    }
    $t \gets t + 1$\;
}
\Return{$\theta(C; \text{final answer})$}\;
\end{algorithm}

\begin{itemize}
    \item \textbf{Summarization frequency:} Instead of summarizing the trajectory every $n=50$ turns, we summarize every $n=25$ turns.
    \item \textbf{Summarization trigger:} Instead of summarizing the trajectory every $n$ turns, we summarize the trajectory when the input length exceeds a threshold $\tau= \{32768, 65536\}$ tokens.
    \item \textbf{Search tool:} We vary the number of top search results $k = \{10, 20\}$.
    \item \textbf{Browse tool:} We vary the maximum length of the scraped content $L = \{3000, 10000, 20000\}$ characters. We also ablate the chunking and scoring strategy. By default, we chunk by natural paragraphs (splitting at newlines) and use ROUGE-L as the similarity metric. We also try using BM25 \citep{robertson2009bm25} as the similarity metric and splitting the content into chunks of 100 words (splitting at any whitespace).
\end{itemize}
For these ablations, we use o4-mini as the base model due to its cheaper cost and test on a smaller subset of 50 samples for each dataset.
The results are shown in \Cref{tab:ab_ours}.

\begin{table}[!th]
    \caption{
        Ablation results with o4-mini as the base model.
        The number of tokens is shown in 10,000s. The cost is shown in US dollars. 
        We ablate design choices in the summarization module, chunking strategy, and search and browse tool.
        For all settings, we set the tool budget to 100.
        The default setting summarizes every $n=50$ turns, chunks by newline, use ROUGE-L as the similarity metric, and search returns the top $k=10$ search results while browsing returns at most $L=10,000$ characters.
        These experiments use a smaller subset of 100 samples for each dataset, so they are not directly comparable to the main results.
        Each experiment is run with three random seeds and the results are the mean and standard deviation.
    }
    \label{tab:ab_ours}
    \centering
    \small
    \vspace{2.0pt}
    \resizebox{0.98\linewidth}{!}{
    \begin{tabular}{lrrrrrrrrrrrrr}
        \toprule
        & & \multicolumn{4}{c}{\textbf{BrowseComp}} & \multicolumn{4}{c}{\textbf{HLE}}  \\
        \cmidrule(lr){3-6} \cmidrule(lr){7-10}
        & & Score ($\uparrow$) & Tokens ($\downarrow$) & Tools ($\downarrow$) & Cost ($\downarrow$) & Score ($\uparrow$) & Tokens ($\downarrow$) & Tools ($\downarrow$) & Cost ($\downarrow$) \\
        \midrule
        \ours & Default & 40.67±5.86 & 118.27±5.93 & 54.02±2.43 & 1.33±0.07 & 17.33±3.06 & 11.39±1.34 & 7.61±0.46 & 0.13±0.01 \\
        \midrule
        \multicolumn{4}{l}{\bf Summarization Module}  \\
        \midrule
        \multicolumn{2}{l}{$n=25$} & 30.33±4.51 & 57.64±4.87 & 35.47±1.04 & 0.65±0.05 & 21.67±4.04 & 8.53±1.68 & 6.09±0.96 & 0.1±0.02 \\
        \multicolumn{2}{l}{Summarize at 32K tokens} & 29.67±2.89 & 46±3.23 & 32.7±0.79 & 0.53±0.04 & 17.67±6.51 & 9.22±1.34 & 6.62±0.69 & 0.11±0.02 \\
        \multicolumn{2}{l}{Summarize at 64K tokens} & 42.67±2.08 & 126.2±5.69 & 57.23±2.14 & 1.42±0.06 & 19.67±4.51 & 11.67±0.74 & 7.83±0.34 & 0.13±0.01 \\
        \midrule
        \multicolumn{4}{l}{\bf Chunking} \\
        \midrule
        \multicolumn{2}{l}{Split newline, BM25} & 37.67±3.51 & 121.38±8.56 & 55.3±2.01 & 1.37±0.1 & 21.33±4.04 & 12.21±1.14 & 8.33±0.31 & 0.14±0.01 \\
        \multicolumn{2}{l}{Split words, ROUGE} & 39.33±5.03 & 113.55±2.88 & 52.97±1.39 & 1.28±0.03 & 19.33±5.69 & 11.69±1.05 & 7.91±0.44 & 0.13±0.01 \\
        \multicolumn{2}{l}{Split words, BM25} & 40.67±2.52 & 121.39±3.33 & 55.95±1.18 & 1.37±0.04 & 20.33±4.16 & 10.4±0.66 & 7.32±0.53 & 0.12±0.01 \\
        \midrule
        \multicolumn{4}{l}{\bf Search and Browse} \\
        \midrule
        \multicolumn{2}{l}{No visit} & 34.33±2.89 & 111.53±5.1 & 63.47±2.03 & 1.26±0.06 & 15.33±1.15 & 14.35±1.68 & 10.42±0.9 & 0.16±0.02 \\
        \multicolumn{2}{l}{No query in visit} & 37.33±1.15 & 187.63±9.54 & 66.82±2.34 & 2.1±0.11 & 20.33±2.08 & 14.55±0.75 & 8.9±0.62 & 0.17±0.01 \\
        \multicolumn{2}{l}{$k=10, L=3,000$} & 42±6.24 & 111.59±12.51 & 52.77±3.95 & 1.26±0.14 & 21.33±1.53 & 11.65±1.2 & 7.75±0.07 & 0.13±0.01 \\
        \multicolumn{2}{l}{$k=10, L=20,000$} & 38.67±1.53 & 117.35±7.79 & 54.5±1.53 & 1.32±0.09 & 20.67±0.58 & 12.19±1.24 & 7.84±0.57 & 0.14±0.01 \\
        \bottomrule
    \end{tabular}
    }
\end{table}

\subsection{Experimental Details}
\label{app:experimental}

We use o3, o4-mini, \gptoss, \glm, \tongyidr, and Claude-4-Sonnet as our base models.
To calculate the cost, we use the prices listed in \Cref{tab:prices}, which are obtained from respective websites \url{https://platform.openai.com/docs/models/o3}, \url{https://platform.openai.com/docs/models/o4-mini}, \url{https://www.together.ai/models/gpt-oss-120b}, \url{https://openrouter.ai/z-ai/glm-4.7-flash}, \url{https://openrouter.ai/alibaba/tongyi-deepresearch-30b-a3b}, \url{https://claude.com/pricing#api}, \url{https://www.firecrawl.dev/pricing}.

For all models, we use a temperature of $1.0$ and a maximum output token of $32,768$.
For o3 and o4-mini, we always use the default reasoning effort of "medium" and for Claude-4-Sonnet, we set the maximum number of thinking tokens to $30,000$.

To calculate the token cost, we take a weighted sum of the token usage across all LLM calls: non-cached input tokens plus the total output tokens, and multiply the results by price per token (which is different for input and output tokens).
We exclude cached tokens from the calculation because in practice, long-horizon systems are expected to have a large amount of cached tokens and system implementation that takes advantage of caching.
Then, for the total cost, we add in the number of search API and scrape URL operations, multiplied by their respective prices.
For the number of tool calls, we count the number of times the search API and scrape operations, the two atomic tool operations, are called.

We also include the results of other trained systems in \Cref{tab:main_results}.
For OpenAI Deep Research (DR), the HLE number from the original blog post\footnote{\url{https://openai.com/index/introducing-deep-research/}} and the BrowseComp number is from the BrowseComp paper \citep{wei2025browsecompsimplechallengingbenchmark}.
For Grok-4, the HLE number is from the original Grok 4 blog post \footnote{\url{https://x.ai/news/grok-4}} and the BrowseComp number is from the Grok 4 Fast blog post \footnote{\url{https://x.ai/news/grok-4-fast}}.
The WebResearcher (WebR) numbers are from the original paper \citep{qiao2025webresearcherunleashingunboundedreasoning}, where we show the results of the main WebResearcher-30B-A3B model; we exclude the heavy version since it uses multiple samples and aggregate the results.
The WebThinker (WebT) numbers are from the original paper \citep{Li2025WebThinker}, where we show the results of the main WebThinker-32B model. They did not evaluate on BrowseComp, so we only report the HLE number.

\begin{table}[h!]
    \caption{
        Pricing for different components.
        Numbers are obtained from respective websites: open-weight models use pricing from OpenRouter.
    }
    \label{tab:prices}
    \centering
    \begin{tabular}{lr}
    \toprule
     & Cost \\
    \midrule
    o3 & \$2.0 / M input token, \$8.0 / M output token \\
    o4-mini & \$1.1 / M input token, \$4.4 / M output token \\
    Claude-4-Sonnet & \$3.0 / M input token, \$15.0 / M output token \\
    GPT-OSS-120B & \$0.15 / M input token, \$0.6 / M output token \\
    Tongyi-DeepResearch & \$0.09 / M input token, \$0.45 / M output token \\
    GLM-4.7 Flash & \$0.07 / M input token, \$0.4 / M output token \\
    Google search & \$0.5 / K query \\
    Scrape URL & \$0.83 / K query \\
    \bottomrule
\end{tabular}
\end{table}

\subsection{Additional Results}
\label{app:additional_results}

\paragraph{Main Results.} We show the results of \ours{} with o3 as the base model over three random seeds in \Cref{tab:stats_o3}. Here we also provide the concrete results for \ours{} with different base models---Claude-4-Sonnet is shown in \Cref{tab:main_results_claude}.
For BrowseComp, HLE, and GAIA, we report the final correctness score. For DeepSearchQA, we report the fully correct score. For HealthBench Hard, we report the overall score.
Due to API costs, we only evaluate Claude-4-Sonnet and \tongyidr{} on BrowseComp and HLE.

We show additional results for \gptoss{} and \glm{} in \Cref{tab:main_results_open_full}.
For \glm, we also test using discard as a context management strategy, where the entire context is discarded when the rollout reaches the summarization step (i.e., 50 turns in our settings).
We find that summarization performs substantially better on BrowseComp and DeepSearchQA and only marginally worse on HLE, GAIA, and HealthBench Hard, thus, we chose to use summarization for the main experiments.
For \gptoss, we also report the results for \ours{} with a tool budget of $T=100$ so the total costs are more comparable with the baselines.

\begin{table}[!th]
    \vspace{-5pt}
    \caption{
        Main results with o3 and o4-mini as the base model.
        All results are macro-averaged across test instances. 
        The number of tokens is shown in 10,000s. The cost is shown in US cents. 
        $T$ denotes the tool budget.
        For reference only, $\dagger$ marks deep research systems that underwent task-specific training.
        Numbers are from the original reports \citep{OpenAIDeepResearch2025,xAI_Grok32025,qiao2025webresearcherunleashingunboundedreasoning,Li2025WebThinker}, and are not directly comparable due to different subsets of test instances used.
    }
    \label{tab:main_results}
    \centering
    \small
    \vspace{2.0pt}
    \resizebox{0.98\linewidth}{!}{

    \begin{NiceTabular}{clc|*{15}{w{c}{0.75cm}}}
        \toprule
        & \Block{2-1}{\textbf{Model}} & \Block{2-1}{\textbf{T}} & \multicolumn{3}{c}{\textbf{BrowseComp}} & \multicolumn{3}{c}{\textbf{HLE}} & \multicolumn{3}{c}{\textbf{DeepSearchQA}} & \multicolumn{3}{c}{\textbf{GAIA}} & \multicolumn{3}{c}{\textbf{HealthBench Hard}} \\
        \cmidrule(lr){4-6}\cmidrule(lr){7-9}\cmidrule(lr){10-12}\cmidrule(lr){13-15}\cmidrule(lr){16-18}
        & & & Tools & Cost & Score & Tools & Cost & Score & Tools & Cost & Score & Tools & Cost & Score & Tools & Cost & Score \\
        \midrule\midrule
        \Block[fill=red!8]{17-1}{\rotatebox{90}{\textbf{O3}}} & Base & - & 0.0 & 7.6 & 17.0 & 0.0 & 5.3 & 18.3 & 0.0 & 2.2 & 20.7 & 0.0 & 2.4 & 35.0 & 0.0 & 0.9 & 35.9 \\
        \cmidrule(lr){2-18}
        & \Block{5-1}{\ours} & 10 & 9.6 & 6.6 & 22.0 & 4.1 & 9.2 & 25.7 & 9.6 & 6.8 & 21.7 & 7.4 & 6.6 & 53.4 & 1.4 & 1.8 & 42.4 \\
        & & 25 & 21.5 & 20.8 & 33.0 & 7.3 & 18.6 & 29.7 & 22.0 & 24.2 & 35.0 & 13.4 & 17.2 & 67.0 & 1.6 & 2.0 & 41.3 \\
        & & 50 & 36.7 & 55.5 & 44.7 & 9.4 & 27.6 & 30.7 & 37.1 & 74.9 & 56.0 & 20.2 & 39.3 & 69.9 & 1.6 & 2.1 & 38.5 \\
        & & 100 & 47.3 & 105.7 & 53.7 & 9.3 & 30.6 & 31.7 & 49.5 & 90.9 & 59.0 & 24.7 & 49.8 & 74.8 & 1.8 & 2.1 & 42.0 \\
        & & 150 & 75.9 & 124.6 & \textbf{56.0} & 11.4 & 36.2 & \textbf{33.3} & 60.5 & 114.1 & \textbf{62.0} & 23.6 & 47.5 & \textbf{77.7} & 1.6 & 2.0 & \textbf{42.5} \\
        \cmidrule(lr){2-18}
        & \Block{3-1}{\react} & 1 & 9.0 & 7.9 & 4.3 & 5.7 & 9.7 & 16.0 & 1.0 & 1.3 & 17.0 & 0.9 & 1.7 & 30.1 & 2.9 & 4.5 & 39.8 \\
        & & 5 & 19.2 & 14.8 & 6.7 & 10.0 & 12.3 & 19.7 & 4.1 & 2.7 & 20.7 & 3.5 & 2.9 & 30.1 & 2.8 & 3.7 & 39.2 \\
        & & 10 & 22.9 & 17.9 & 7.0 & 10.9 & 14.9 & 21.3 & 6.9 & 4.6 & 17.3 & 6.0 & 4.4 & 35.9 & 3.0 & 4.2 & 37.1 \\
        \cmidrule(lr){2-18}
        & \Block{6-1}{\searcho} & 1 & 9.5 & 8.8 & 18.0 & 5.2 & 7.5 & 20.0 & 10.1 & 5.4 & 28.7 & 8.5 & 5.0 & 50.5 & 1.2 & 1.4 & 39.2 \\
        & & 5 & 46.9 & 22.2 & 24.0 & 18.7 & 13.7 & 20.7 & 46.1 & 17.7 & 32.3 & 28.4 & 11.8 & 65.0 & 2.2 & 1.8 & 37.8 \\
        & & 10 & 89.8 & 39.2 & 31.0 & 23.9 & 16.7 & 26.3 & 84.1 & 32.6 & 35.0 & 45.8 & 18.6 & 63.1 & 2.5 & 1.8 & 37.6 \\
        & & 25 & 183.2 & 79.8 & 40.0 & 44.2 & 28.5 & 25.0 & 169.9 & 70.1 & 40.0 & 79.0 & 34.8 & 67.0 & 2.2 & 1.8 & 39.5 \\
        & & 50 & 306.2 & 143.3 & 48.3 & 49.8 & 32.6 & 27.0 & 248.2 & 113.1 & 45.3 & 109.9 & 52.8 & 67.0 & 2.6 & 1.9 & 37.4 \\
        & & 100 & 456.7 & 246.7 & 55.7 & 52.2 & 37.1 & 27.0 & 353.5 & 189.7 & 44.0 & 141.7 & 79.1 & 69.9 & 2.5 & 1.8 & 38.5 \\
        \cmidrule(lr){2-18}
        & \hfodr & - & 8.4 & 59.5 & 20.0 & 1.7 & 18.5 & 17.7 & 6.8 & 52.8 & 30.7 & 5.1 & 45.2 & 51.5 & 1.4 & 6.2 & 32.1 \\
        \cmidrule(lr){2-18}
        & \gptoss & - & 69.5 & 29.3 & 10.7 & 85.6 & 31.9 & 16.0 & 47.6 & 20.2 & 23.3 & 48.8 & 21.0 & 45.6 & 104.6 & 38.7 & 16.3 \\
        \midrule
        \Block[fill=blue!8]{17-1}{\rotatebox{90}{\textbf{O4-Mini}}} & Base & - & 0.0 & 5.7 & 5.0 & 0.0 & 2.5 & 15.0 & 0.0 & 1.6 & 14.0 & 0.0 & 1.8 & 28.2 & 0.0 & 0.5 & 18.2 \\
        \cmidrule(lr){2-18}
        & \Block{5-1}{\ours} & 10 & 9.8 & 7.7 & 12.7 & 3.4 & 4.8 & 20.7 & 9.7 & 5.4 & 19.7 & 7.2 & 4.5 & 50.5 & 1.5 & 1.1 & 23.0 \\
        & & 25 & 22.8 & 27.4 & 21.3 & 5.4 & 8.0 & 24.0 & 22.6 & 21.1 & 24.3 & 14.6 & 14.5 & 59.2 & 2.5 & 1.9 & 22.9 \\
        & & 50 & 40.3 & 84.8 & 31.0 & 7.1 & 12.2 & 25.7 & 40.1 & 65.2 & 37.7 & 20.5 & 29.0 & 65.0 & 2.6 & 2.2 & 22.7 \\
        & & 100 & 48.5 & 116.8 & 33.3 & 7.2 & 12.9 & \textbf{26.7} & 60.6 & 94.4 & 41.3 & 23.5 & 32.3 & \textbf{71.8} & 2.4 & 2.2 & 22.3 \\
        & & 150 & 84.1 & 169.5 & \textbf{37.7} & 7.8 & 13.7 & 25.3 & 73.8 & 116.8 & \textbf{46.3} & 25.7 & 36.8 & 67.0 & 2.4 & 1.8 & 22.4 \\
        \cmidrule(lr){2-18}
        & \Block{3-1}{\react} & 1 & 9.3 & 5.8 & 1.3 & 4.6 & 4.8 & 17.0 & 1.0 & 0.6 & 7.0 & 0.8 & 0.9 & 21.4 & 1.6 & 1.6 & 22.1 \\
        & & 5 & 18.9 & 10.0 & 3.0 & 6.4 & 5.5 & 15.3 & 4.7 & 2.2 & 10.7 & 3.8 & 2.3 & 29.1 & 2.1 & 1.9 & 20.8 \\
        & & 10 & 20.6 & 9.8 & 2.3 & 6.9 & 5.9 & 15.3 & 7.5 & 3.8 & 12.0 & 5.7 & 3.3 & 32.0 & 2.4 & 1.9 & 21.6 \\
        \cmidrule(lr){2-18}
        & \Block{6-1}{\searcho} & 1 & 10.0 & 8.1 & 6.3 & 3.5 & 3.4 & 13.0 & 9.7 & 3.8 & 15.7 & 8.0 & 3.4 & 42.7 & 0.7 & 0.7 & 20.1 \\
        & & 5 & 49.7 & 21.3 & 11.3 & 11.9 & 6.1 & 23.3 & 45.2 & 13.9 & 27.3 & 29.9 & 9.1 & 57.3 & 1.8 & 1.0 & 20.8 \\
        & & 10 & 93.9 & 36.3 & 17.3 & 15.6 & 7.3 & 17.0 & 79.8 & 25.4 & 31.3 & 41.9 & 13.7 & 54.4 & 1.7 & 1.0 & \textbf{23.2} \\
        & & 25 & 207.7 & 74.8 & 25.0 & 22.5 & 9.2 & 22.3 & 130.6 & 40.4 & 32.3 & 22.2 & 12.2 & 34.0 & 2.7 & 1.2 & 20.5 \\
        & & 50 & 351.5 & 126.0 & 28.7 & 26.3 & 11.6 & 19.3 & 226.3 & 70.2 & 35.7 & 82.1 & 24.3 & 59.2 & 1.7 & 0.9 & 21.6 \\
        & & 100 & 546.7 & 201.6 & 36.0 & 25.8 & 10.8 & 21.3 & 279.4 & 89.2 & 34.0 & 83.6 & 25.9 & 55.3 & 2.0 & 1.1 & 19.6 \\
        \cmidrule(lr){2-18}
        & \hfodr & - & 15.4 & 52.1 & 15.0 & 3.9 & 12.8 & 16.3 & 17.1 & 77.0 & 34.7 & 11.7 & 49.2 & 62.1 & 6.0 & 19.9 & -0.1 \\
        \cmidrule(lr){2-18}
        & \gptr & - & 82.5 & 29.5 & 4.0 & 100.8 & 32.8 & 11.3 & 56.3 & 22.6 & 15.7 & 62.9 & 23.1 & 42.7 & 112.3 & 35.6 & -0.2 \\
        \midrule
        \Block[fill=gray!3]{4-1}{\rotatebox{90}{\textbf{Others}}} & \headercolor OpenAI DR$^\dagger$ & - & - &- & 51.5 & - & - &  26.6 & - & - & -  \\
         & \headercolor Grok-4$^\dagger$ & - & - & - & 43.0 & - & - & 38.6 & - & - & -  \\
         & \headercolor WebR-30B$^\dagger$ & - & - & - & 37.3 &  - & - & 28.8 & - & - & -  \\
         & \headercolor WebT-32B$^\dagger$ & - & - & - & 15.8 &  - & - & - & - & - & -  \\
        \bottomrule
    \end{NiceTabular}

    }

    \vspace{-5pt}
\end{table}

\begin{table}[tbp]
    \centering
    \resizebox{\linewidth}{!}{%

    \begin{NiceTabular}{clc|*{15}{w{c}{0.75cm}}}
        \toprule
        & \Block{2-1}{\textbf{Model}} & \Block{2-1}{\textbf{T}} & \multicolumn{3}{c}{\textbf{BrowseComp}} & \multicolumn{3}{c}{\textbf{HLE}} & \multicolumn{3}{c}{\textbf{DeepSearchQA}} & \multicolumn{3}{c}{\textbf{GAIA}} & \multicolumn{3}{c}{\textbf{HealthBench Hard}} \\
        \cmidrule(lr){4-6}\cmidrule(lr){7-9}\cmidrule(lr){10-12}\cmidrule(lr){13-15}\cmidrule(lr){16-18}
        & & & Tools & Cost & Score & Tools & Cost & Score & Tools & Cost & Score & Tools & Cost & Score & Tools & Cost & Score \\
        \midrule\midrule
        \Block[fill=red!8]{5-1}{\rotatebox{90}{\textbf{GLM-4.7}}} & Base & - & 0.0 & 0.7 & 1.3 & 0.0 & 0.7 & 9.7 & 0.0 & 0.4 & 7.3 & 0.0 & 0.5 & 14.6 & 0.0 & 0.1 & 6.8 \\
        & \react & 10 & 27.8 & 3.0 & 2.0 & 17.8 & 2.2 & 9.0 & 30.1 & 2.9 & 8.0 & 24.7 & 2.6 & 18.4 & 15.6 & 1.5 & 12.9 \\
        & \searcho & 50 & 155.3 & 15.9 & 7.7 & 49.4 & 5.3 & 9.7 & 116.0 & 11.7 & 19.0 & 69.0 & 7.0 & 45.6 & 23.0 & 2.3 & 11.5 \\
        & \ours & 150 & 51.8 & 4.9 & \textbf{17.3} & 19.8 & 2.5 & 13.7 & 48.8 & 5.1 & \textbf{20.7} & 26.0 & 2.7 & 49.5 & 5.2 & 0.5 & 13.7 \\
        & \ours-Discard & 150 & 51.5 & 4.6 & 14.0 & 19.1 & 2.3 & \textbf{14.7} & 53.1 & 5.6 & 14.3 & 27.5 & 2.9 & \textbf{50.5} & 5.0 & 0.4 & \textbf{14.5} \\
        \midrule
        \Block[fill=blue!8]{5-1}{\rotatebox{90}{\textbf{GPT-OSS}}} & Base & - & 0.0 & 0.2 & 2.7 & 0.0 & 0.2 & 10.0 & 0.0 & 0.1 & 9.3 & 0.0 & 0.1 & 23.3 & 0.0 & 0.1 & 36.6 \\
        & \react & 10 & 26.1 & 3.2 & 3.3 & 10.3 & 1.4 & 11.0 & 34.8 & 4.3 & 4.7 & 25.0 & 3.1 & 30.1 & 4.8 & 0.8 & 39.1 \\
        & \searcho & 50 & 45.7 & 4.4 & 6.0 & 9.7 & 1.1 & 16.0 & 60.1 & 5.8 & 15.0 & 29.1 & 2.9 & 45.6 & 2.6 & 0.4 & 37.6 \\
        & \Block{2-1}{\ours} & 100 & 60.5 & 10.6 & \textbf{20.0} & 7.8 & 1.3 & 17.0 & 51.3 & 9.8 & 34.0 & 19.6 & 3.0 & 55.3 & 2.2 & 0.4 & 38.6 \\
        & & 150 & 70.5 & 12.5 & 20.0 & 8.1 & 1.3 & \textbf{19.0} & 61.6 & 11.6 & \textbf{34.3} & 23.5 & 3.9 & \textbf{58.3} & 2.2 & 0.4 & \textbf{39.3} \\
        \bottomrule
    \end{NiceTabular}

    }
    \caption{Additional results for \gptoss{} and \glm{}. Costs are in US cents. For \glm, we also test using discard as a context management strategy (\ours-Discard). For \gptoss, we also report the results for \ours{} with a tool budget of $T=100$ so the total costs are more comparable with the baselines.}
    \label{tab:main_results_open_full}
\end{table}

\begin{table}[!th]
    \caption{
        Statistical significance analysis with o3 as the base model. We run with three random seeds for each experiment and report the mean and standard deviation.
    }
    \label{tab:stats_o3}
    \centering
    \small
    \vspace{2.0pt}
    \resizebox{0.98\linewidth}{!}{
    \begin{tabular}{lrrrrrrrrrrrrr}
        \toprule
        & & \multicolumn{4}{c}{\textbf{BrowseComp}} & \multicolumn{4}{c}{\textbf{HLE}}  \\
        \cmidrule(lr){3-6} \cmidrule(lr){7-10}
        & & Score ($\uparrow$) & Tokens ($\downarrow$) & Tools ($\downarrow$) & Cost ($\downarrow$) & Score ($\uparrow$) & Tokens ($\downarrow$) & Tools ($\downarrow$) & Cost ($\downarrow$) \\
        \midrule
        \headercolor o3 & - & 17.22±1.02 & 3.87±0.12 & 0±0 & 0.08±0 & 19.56±1.07 & 2.63±0.04 & 0±0 & 0.05±0 \\
        Search-o1 & 50 & 49.33±1.2 & \textbf{49.98±1.5} & 298.9±6.84 & 1.24±0.04 & 26.78±0.69 & \textbf{13.05±0.52} & 50.96±1.17 & \textbf{0.3±0.01} \\
        SLIM & 150 & \textbf{53±1.2} & 54.77±5.23 & \textbf{50.84±0.44} & \textbf{1.12±0.1} & \textbf{32.11±1.84} & 16.44±1.15 & \textbf{10.3±0.98} & 0.33±0.02 \\
        \bottomrule
    \end{tabular}
    }
\end{table}

\begin{figure}[t!]
    \centering
    \includegraphics[width=0.95\linewidth]{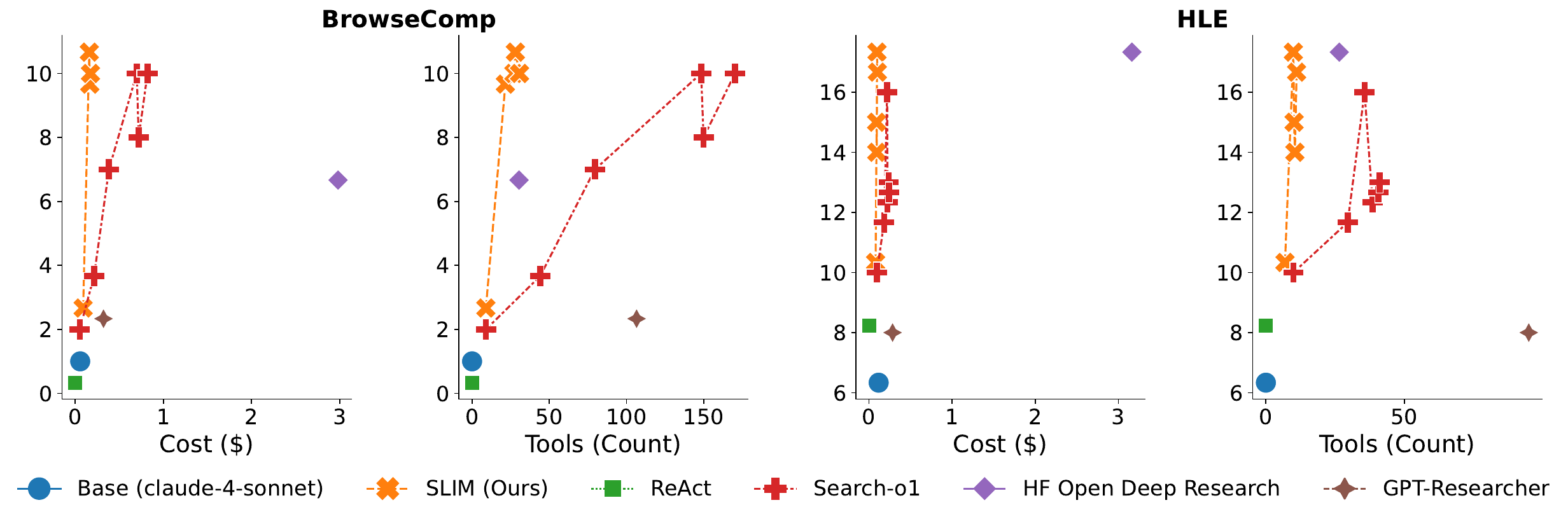}
    \vspace{-4.0pt}
    \caption{
        With Claude-4-Sonnet as the base model, \ours{} consistently outperforms other baselines on BrowseComp while using fewer tool calls and lower overall costs. On HLE, \ours{} can achieve overall higher performance and use fewer tool calls.
    }
    \label{fig:main_claude}
\end{figure}

\begin{table}[!th]
    \caption{
        Main results with Claude-4-Sonnet as the base model.
        All results are macro-averaged across test instances. 
        The number of tokens is shown in 10,000s. The cost is shown in US dollars. 
        $T$ denotes the maximum number of turns in each trajectory.
    }
    \label{tab:main_results_claude}
    \centering
    \small
    \vspace{2.0pt}
    \resizebox{0.98\linewidth}{!}{
    \begin{tabular}{lrrrrrrrrrrrrr}
        \toprule
        & & \multicolumn{4}{c}{\textbf{BrowseComp}} & \multicolumn{4}{c}{\textbf{HLE}}  \\
        \cmidrule(lr){3-6} \cmidrule(lr){7-10}
        & $T$ & Score ($\uparrow$) & Tokens ($\downarrow$) & Tools ($\downarrow$) & Cost ($\downarrow$) & Score ($\uparrow$) & Tokens ($\downarrow$) & Tools ($\downarrow$) & Cost ($\downarrow$) \\
        \midrule

         Claude-4-Sonnet & - & 1.0 & 1.9 & 0.0 & 0.06 & 6.3 & 3.9 & 0.0 & 0.12 \\
         \midrule
        \multirow{3}{*}{\react} & 1 & 0.3 & 0.0 & 0.0 & 0.00 & 8.3 & 0.0 & 0.0 & 0.00 \\
        & 5 & 0.3 & 0.0 & 0.0 & 0.00 & 8.3 & 0.0 & 0.0 & 0.00 \\
        & 10 & 0.3 & 0.0 & 0.0 & 0.00 & 8.0 & 0.0 & 0.0 & 0.00 \\
        \midrule
        \multirow{6}{*}{\searcho} & 1 & 2.0 & 1.5 & 9.0 & 0.05 & 10.0 & 2.9 & 10.0 & 0.09 \\
        & 5 & 3.7 & 6.0 & 44.1 & 0.21 & 11.7 & 5.3 & 29.5 & 0.18 \\
        & 10 & 7.0 & 10.7 & 79.5 & 0.38 & 16.0 & 6.3 & 35.6 & 0.22 \\
        & 25 & 8.0 & 20.1 & 149.9 & 0.72 & 13.0 & 6.8 & 41.1 & 0.24 \\ 
        & 50 & 10.0 & 22.9 & 170.3 & 0.82 & 12.7 & 7.0 & 40.7 & 0.24 \\
        & 100 & 10.0 & 19.4 & 148.3 & 0.70 & 12.3 & 6.4 & 38.5 & 0.22 \\
        \midrule
        \hfodr & 20 & 6.7 & 98.8 & 30.4 & 2.98 & 17.3 & 105.0 & 26.5 & 3.16 \\
        \gptr & - & 2.3 & 7.9 & 106.5 & 0.32 & 8.0 & 6.9 & 94.9 & 0.28 \\
        \midrule
        \multirow{5}{*}{\ours} & 10 & 2.7 & 2.8 & 8.9 & 0.09 & 10.3 & 2.5 & 6.9 & 0.08 \\
        & 25 & 9.7 & 5.1 & 21.6 & 0.17 & 15.0 & 2.8 & 10.2 & 0.09 \\
        & 50 & 10.0 & 5.0 & 27.1 & 0.16 & 17.3 & 3.0 & 9.9 & 0.10 \\
        & 100 & \textbf{10.7} & 4.8 & 28.1 & 0.16 & 14.0 & 2.9 & 10.5 & 0.09 \\
        & 150 & 10.0 & 5.2 & 30.7 & 0.17 & \textbf{16.7} & 3.1 & 11.1 & 0.10 \\

        \bottomrule
    \end{tabular}
    }
\end{table}

\paragraph{\react{} Ablations.}
We vary the number of search results $k$ and the maximum length of the scraped content $L$ for \react{} to see the effect of search tool design choices, as shown in \Cref{tab:ab_react}.
We found that overall there aren't significant differences in the HLE results, but using fewer search results $k=5$ than the default $k=10$ leads to a 2.7 points improvement in the BrowseComp results.
This is likely due to the fact that search results lower in the ranking are often noisy and irrelevant to the question, and using fewer but more relevant search results leads to a more focused search process.
Furthermore, fewer search results means less context is added to the LLM, preventing it from hitting the context window limit as much. This is evident in more token and tool usage.
However, we use $k=10$ for the main experiments to stay consistent with the other baselines.

\begin{table}[!th]
    \vspace{-5pt}
    \caption{
        \react{} ablations with o3 as the base model, and the maximum number of turns is $T = 10$.
        We vary the number of search results $k$ and the maximum length of the scraped content $L$.
    }
    \label{tab:ab_react}
    \centering
    \small
    \vspace{2.0pt}
    \resizebox{0.98\linewidth}{!}{
    \begin{tabular}{lrrrrrrrrrrrrrrr}
        \toprule
        & \multicolumn{3}{c}{\bf Parameters} & \multicolumn{4}{c}{\textbf{BrowseComp}} & \multicolumn{4}{c}{\textbf{HLE}}  \\
        \cmidrule(lr){2-4} \cmidrule(lr){5-8} \cmidrule(lr){9-12}
        & $T$ & $k$ & $L$ & Score ($\uparrow$) & Tokens ($\downarrow$) & Tools ($\downarrow$) & Cost ($\downarrow$) & Score ($\uparrow$) & Tokens ($\downarrow$) & Tools ($\downarrow$) & Cost ($\downarrow$) \\
        \midrule
        \react & 10 & 10 & 10k  & 7.0 & 8.0 & 2.8 & 0.16 & 21.3 & 7.0 & 1.2 & 0.14 \\
        \react & 10 & 5 & 10k  & 9.7 & 10.6 & 4.1 & 0.21 & 21.7 & 7.0 & 1.7 & 0.14 \\
        \react & 10 & 10 & 3k  & 5.0 & 8.7 & 2.8 & 0.18 & 22.7 & 6.5 & 1.2 & 0.13 \\
        \react & 10 & 5 & 3k  & 8.3 & 10.7 & 4.1 & 0.22 & 21.3 & 6.7 & 1.7 & 0.13 \\
        \bottomrule
    \end{tabular}
    }
    \vspace{-5pt}
\end{table}

\subsection{Additional Analysis}
\label{app:additional_analysis}

In this subsection, we provide additional analysis---we extend the initial outcome-based analysis to \ours, and show the trajectory-level analysis on the more comprehensive baselines.

In \Cref{tab:analysis_errors_alt}, we show the trajectory-level analysis where we report the failure modes as a percentage of trajectories that ends with an incorrect answer.
The trends are consistent with the analysis in the main text, but we find that \ours{} can often find the correct answer across its long trajectories---over 69\% of the incorrect trajectories encounters the correct answer, but the model is not able to identify and use it to answer the question.
This could be attributed to the fact that modern LLMs still struggle at long-context settings where it may need to reason over many sources. We leave these improvements to future work.

\begin{table}[!t]
    \vspace{-1em}
    \caption{
        For correct, we report the percentage of trajectories across all samples.
        For each trajectory-level failure mode, we report the percentage of trajectories that ends with an incorrect answer.
        For hallucination only, we report the percentage of hallucinations for samples that ends with an incorrect answer and do not abstain.
    }
    \label{tab:analysis_errors_alt}
    \centering
    \small
    \resizebox{0.98\linewidth}{!}{
    \begin{tabular}{lrrrrrrrr}
        \toprule
        &\multirow{2}{4em}{\raggedleft \textbf{Turn Budget}}& \multirow{2}{4em}{\raggedleft \textbf{Correct}} & \multirow{2}{4em}{\raggedleft \textbf{Confirm Bias}} & \multirow{2}{4em}{\raggedleft \textbf{Unfocused Search}} & \multirow{2}{4em}{\raggedleft \textbf{Inefficient Search}} & \multirow{2}{4em}{\raggedleft \textbf{Abstention}} & \multirow{2}{4em}{\raggedleft \textbf{Answer Ignored}} & \multirow{2}{4em}{\raggedleft \textbf{Hallucinate}} \\
        \textbf{Framework} & & & & & & & \\
        \midrule
        \react & 10 & 7.0 & 10.0 & 47.3 & 4.2 & 1.1 & 0.7 & 56.7 \\
        \searcho & 50 & 48.3 & 18.1 & 65.2 & 14.0 & 8.4 & 50.3 & 46.8 \\
        \hfodr & 20 & 20.0 & 8.6 & 75.5 & 56.5 & 41.6 & 2.1 & 96.2 \\
        \ours & 150 & 56.0 & 22.0 & 77.3 & 17.2 & 62.9 & 69.7 & 19.0 \\

        \bottomrule
    \end{tabular}
    }

\end{table}

\end{document}